\pdfoutput=1

\documentclass[11pt]{article}

\usepackage[preprint]{acl}

\usepackage{times}
\usepackage{latexsym}
\usepackage{times}
\usepackage{latexsym}
\usepackage{hyperref}
\usepackage{color, colortbl}
\definecolor{Gray}{gray}{0.9}
\usepackage{array}
\usepackage{booktabs}
\usepackage{multirow}
\usepackage{amsmath}
\usepackage{amssymb}
\usepackage{graphicx}
\usepackage{enumerate}
\usepackage{diagbox}
\usepackage{mathtools}
\usepackage{paralist}
\usepackage{tabularx}
\usepackage{lipsum}
\usepackage{ragged2e}
\usepackage{makecell}
\usepackage{amsfonts}
\usepackage{amsmath}
\usepackage[shortlabels]{enumitem}

\usepackage[T1]{fontenc}

\usepackage[utf8]{inputenc}
\usepackage{csquotes}

\usepackage{microtype}

\usepackage{inconsolata}

\usepackage{graphicx}

\definecolor{skyblue}{rgb}{0, 0.6, 0.95}

\definecolor{headcolor}{HTML}{018161}
\definecolor{relationcolor}{HTML}{d95f02}
\definecolor{tailcolor}{HTML}{6560a3}

\newcommand{\dataset}{\textsc{SessionIntentBench}}
\newcommand{\taskI}{Intent-Based Purchasing Likelihood Estimation}
\newcommand{\taskII}{Purchasing Likelihood Inference via Valued Attributes Regularization}
\newcommand{\taskIII}{Intention Justification via Comparison}
\newcommand{\taskIV}{Intention Evolution Modeling}

\title{\dataset{}: A Multi-task Inter-session Intention-shift Modeling Benchmark for E-commerce Customer Behavior Understanding}

\author{Yuqi Yang\thanks{Equal Contribution}$^{\spadesuit}$,
Weiqi Wang$^{*\spadesuit\clubsuit}$\thanks{Work done during his internship at Amazon.com Inc.},
Baixuan Xu$^{\spadesuit}$,
Wei Fan$^{\spadesuit}$,
Qing Zong$^{\spadesuit}$,
Chunkit Chan$^{\spadesuit}$,\\
\textbf{
Zheye Deng$^{\spadesuit}$,
Xin Liu$^{\clubsuit}$,
Yifan Gao$^{\clubsuit}$,
Changlong Yu$^{\clubsuit}$,
Chen Luo$^{\clubsuit}$,}\\
\textbf{
Yang Li$^{\clubsuit}$,
Zheng Li$^{\clubsuit}$,
Qingyu Yin$^{\clubsuit}$,
Bing Yin$^{\clubsuit}$,
Yangqiu Song$^{\spadesuit\clubsuit}$\thanks{Visiting academic scholar at Amazon.com Inc.}}\\
$^{\spadesuit}$Department of Computer Science and Engineering, HKUST, Hong Kong SAR, China\\
$^{\clubsuit}$Amazon.com Inc, Palo Alto, CA, USA\\
\texttt{yyangfd@connect.ust.hk; wwangbw@cse.ust.hk; yqsong@cse.ust.hk}\\
}

\begin{document}
\maketitle
\begin{abstract}
Session history is a common way of recording user interaction behaviors throughout a browsing activity involving multiple products.
For example, if a user clicks on a product webpage and then leaves, it might be because certain features do not satisfy the user, which serves as an important indicator of on-the-spot user preferences. 
However, prior works fail to capture and model customer intention effectively because of insufficient information exploitation, relying only on apparent information such as descriptions and titles.
There is also a lack of data and corresponding benchmarks for explicitly modeling intention in E-commerce product purchase sessions.
To address these issues, we introduce the concept of an \textit{intention tree} and propose a dataset curation pipeline. 
Together, we construct a sibling multimodal benchmark,~\dataset{}, that evaluates L(V)LMs' capability to understand inter-session intention shifts through four subtasks.
With 1,952,177 intention entries, 1,132,145 session intention trajectories, and 13,003,664 available tasks mined from 10,905 sessions, we provide a scalable way to exploit existing session data for customer intention understanding. 
We conduct human annotations to collect ground-truth labels for a subset of the collected data to form an evaluation gold set.
Extensive experiments on the annotated data further confirm that current L(V)LMs fail to capture and utilize intention across complex session settings. 
Further analysis shows that injecting intention enhances LLM performance.
\end{abstract}


\section{Introduction}
\begin{figure}[t!]
    \centering
    \includegraphics[width=1\linewidth]{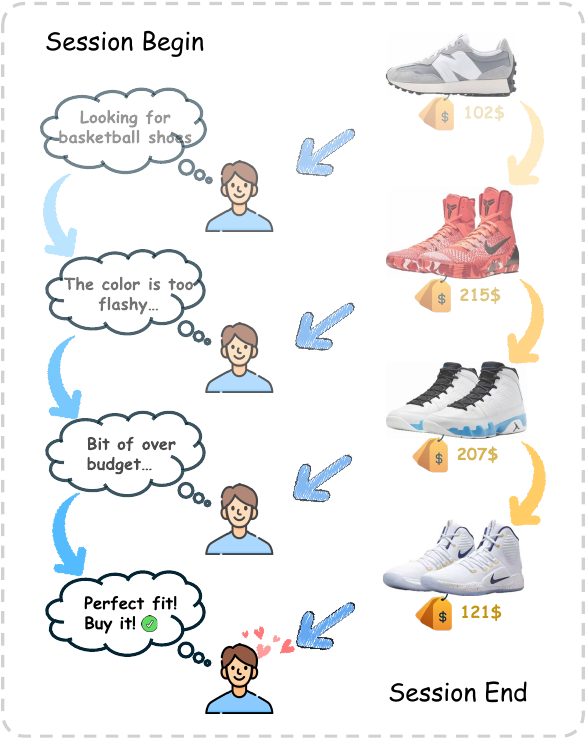}
    \caption{An example of intention shift within a shopping session. As the user interacts with successive products, latent preferences become more specific and can change from one step to the next.}
    \label{figs:introduction}
\end{figure}

Modeling and analyzing customer intention is of great importance in the E-commerce domain~\cite{Dai-2006-commercial-intention,intro_intention_2,intro_intention_3}. 
This enables us to give better product recommendations and provide more personalized services~\cite{intro_personalized_1, intro_personalized_2, intro_personalized_3}. 
Conventional ways of understanding user intention always rely on analyzing user profiles or purchasing records, but such information is not easily retrievable or even missing in real world applications. 
Therefore, we need a data source with better accessibility and applicability, such as the product purchase sessions, which concludes the user behavior throughout a series of sequential browsing activities. 
By analyzing the interaction history in this short period of time, we are able to infer the user intention and how it changes over time. 
The shifting intent behind product searches and inspections can further affect future user interactions. 
For example, in Figure \ref{figs:introduction}, the customer exposes his intention when he switches from flashy red shoes to plain white ones. After that, browsing for shoes at a much lower price shows customers' need for cheap and cheerful products. By modeling customer session intention and adjusting inferred results when needed, we can provide more customized services in an accurate and timely manner.

Existing work either covers session or intention, but not collectively. 
There has been an experiment focusing on exploiting the product information within one session and using it to make direct predictions~\cite{Amazon-M2}, which assembles useful information based on specific product attributes like titles and prices. 
While some other works explicitly model the user intention behind the single purchase or co-buy behaviors \cite{MIND, ding2024-intentionQA, DBLP:conf/eacl/BaiWCYHWLLZLS26}. They leverage the most recent user actions for intention understanding and inference, covering only one or two products, but fall short of exploring user preference shifts over a longer horizon, such as sessions.
However, \citet{Amazon-M2} have shown that session information and fine-grained attribute analysis would help LLMs to give better next-product recommendations. Considering these aspects, it is essential to formulate a method to explicitly model intention over a session period.

But when modeling intention dynamically in more complex purchase contexts, such as sessions, several gaps remain. 
Firstly, current works only use short-term information and focus on single or co-buy purchases. 
This approach overlooks the potential motivational intention embedded in earlier user interactions, therefore hindering the models' capability of making reasonable inferences. 
Furthermore, among various attributes, only product titles and images are used as product inference hints, which omits important dimensions of product information and results in a waste of information from the collected knowledge base. 
Last but not least, we lack an automated pipeline to streamline the construction of such intention data, there hasn't been any formulation of such tasks or benchmark data to evaluate L(V)LM systems.

To combat this, we first propose \dataset{} tasks, consisting of four sequential subtasks tailored to systematically evaluate L(V)LMs' capability in understanding customer intention within session browsing records.
Then, we design an automated framework to streamline the collection of detailed product metadata, customer intention, and intention shift within the session by prompting L(V)LM in a multi-step manner.

By applying our method to Amazon-M2~\cite{Amazon-M2}, we first filter and collect 10,905 sessions with complete textual and visual data. 
We enrich the original session with intention entries and obtain 1,132,145 possible intention pathways.
After that, we further conduct human annotations to 8,980 sampled intention trajectories to form an evaluation benchmark.
Then, we carry out extensive experiments over more than 20 L(V)LMs by applying different evaluation settings and prompting techniques, along with extra fine-tunings.
Our findings indicate that current L(V)LMs struggle with the proposed tasks. 
Further analyses reveal potential underlying causes behind the observed low model accuracy and introduce intention injection as a possible way of assisting models' understanding of session intent and improving performances.

\section{Related Works}
\subsection{Intention Understanding}
Intention is the internal mental state that affects people's decision-making \cite{2002-intent-history}. 
By analyzing the inner intention states of the users, service providers are able to present more personalized products \cite{Dai-2006-commercial-intention} and give back more accurate responses \cite{Zhang-2016-medical-intention}. In E-commerce, customer intention is crucial in understanding their purchase behaviors and preferences \cite{2001-online-purchase-intention}.
There has been ongoing research trying to decode how to model shopping intention. For example, using history information like tags \cite{2025intention} and co-buy behaviors \cite{2023FolkScope, MIND, DBLP:conf/acl/WangCLNX0SLGLYB25}.
Recently, studies show that LLMs are struggling to connect the dots between intended products and user intention \cite{ding2024-intentionQA}. However, figuring out the items the user wanted is even more difficult when it comes to more complex settings like session histories.
To bridge the gap between understanding intention and providing more precise shopping aids, we formulate \dataset{} tasking L(V)LMs to infer intent by leveraging session metadata from multiple angles.

\subsection{Purchase Session in E-commerce}
Purchase session is a record of customer interaction history, which has been becoming an increasingly hot area of research \cite{2022session-3,2023session-1,2024session-2}. Various methods are proposed trying to exploit the abundant information contained here, such as using deep reinforcement learning models \cite{2022sessionintent-deepRL}, leveraging graph neural networks \cite{2023sessionintent-GNN}, and carrying out complex logical reasoning techniques \cite{liuxin2023sessionintent}.
While \citet{Amazon-M2} systematically introduces session information as an important factor for understanding sequential interacting behavior, \citet{liuxin2023sessionintent} points out that product attributes play a pivotal role in enhancing user intent capture. This shows that a more fine-grained framework of session intention evaluation is needed. 
Furthermore, recognizing that multiple intentions can coexist within a session, researchers have explored various approaches to enhance product recommendations. \citet{extra1-LLM} iteratively updates an intention ranking prompt to optimize recommendations, while \citet{extra2-embedding} train a neural network to learn intention embedding representations and refine selections accordingly. While these works aim to provide more precise product recommendation, our research focuses on improving language models' intention understanding and reasoning ability using semantic intention representation.
Using the summarization and generation ability of L(V)LMs, in \dataset{}, we extract and incorporate session intent metadata from multiple aspects for more comprehensive intention capturing.

\section{Problem Definition}
\label{sec:task_definitions}
\subsection{\dataset{} Task Definitions}
\begin{figure*}[t]
    \centering
    \includegraphics[width=1\linewidth]{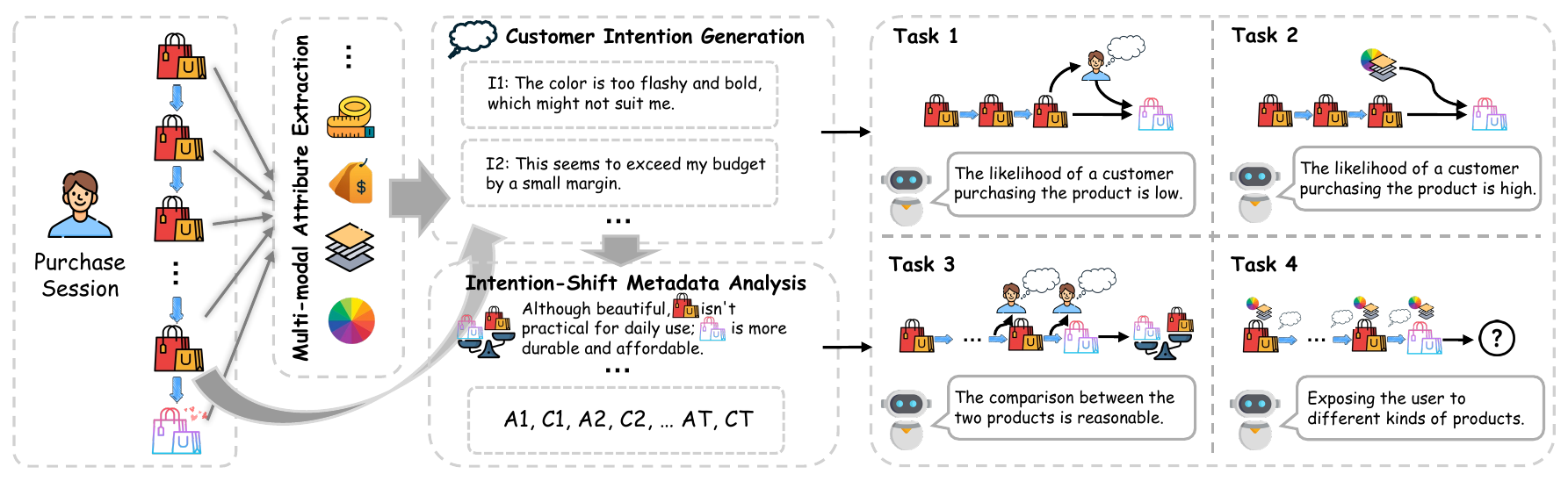}
    \caption{Overview of \dataset{} and the curation pipeline. We first extract multimodal product attributes, then generate candidate intentions and their branches, and finally enrich each branch with attribute and comparison metadata for the four downstream tasks. Here, $A_i$ and $C_i$ denote the valued attribute and comparison metadata at step $i$.}
    \label{fig:maingraph}
\end{figure*}

We use \emph{intention shift} to refer to the step-to-step evolution of user preference across successive interactions in a single shopping session. In the rest of the paper, we use \emph{intention} and \emph{intent} interchangeably. Figure~\ref{fig:maingraph} summarizes our formulation.

We propose to model the intention shift from four aspects, as outlined in Figure~\ref{fig:maingraph}, to facilitate the creation of a L(V)LM shopping agent that is able to:
(i) Detect the attribute that is decisive in the intention shift. 
(ii) Model intention trajectories with mined attributes and leverage them to give better predictions on future interactions. 
(iii) Compare between the most recently viewed product with previously interacted ones and use this comparison to validate the plausibility of the inferred intent.
(iv) Leverage modeled intention trajectories to predict future product interaction preferences. 

Formally, assume a session contains products $P_1,P_2,\ldots,P_T$ observed in chronological order. Let $A_t$ denote the valued attribute associated with the transition at step $t$, $I_t$ the inferred user intention after interacting with $P_t$, and $C_t$ a comparison between adjacent products that helps justify the shift from the previous state to the current one. The interaction history up to time step $t$ is
\[
\mathcal{H}_t=\{(P_j,A_j)\}_{j=1}^{t}.
\]
Our goal is to evaluate whether a model can use $\mathcal{H}_t$ and the mined intention metadata to reason about future interactions and the trajectory of the session.
\noindent\textbf{\\[4px] \textsc{Task} 1: \taskI}. The first task asks the model to verify whether the last proposed intention is well aligned with the new product we are going to interact with. 
The model will be given historical information $\mathcal{H}_{t-1}$, the proposed intention $I_{t-1}$, and the new product $P_t$. It is asked to output a likelihood estimation score $\mathcal{S}_1(P_t, I_{t-1})\mid_{\mathcal{H}_{t-1}}$ $\in$ $\{0, 1, 2, 3\}$ for the customer to interact with $P_t$, where 3 means the most likely and 0 means the least probable.
%
%
%
\noindent\textbf{\\[4px] \textsc{Task} 2: \taskII}. The second task requires the model to verify whether the proposed valued attributes of the user are essential elements of the actual unseen product.
The model is provided with historical information $\mathcal{H}_{t-1}$, the proposed valued attribute $A_{t-1}$, and the new unseen product $P_t$. The model is required to output an estimated interaction likelihood score $\mathcal{S}_2(P_t, A_{t-1})\mid_{\mathcal{H}_{t-1}}$ $\in$ $\{0, 1, 2, 3\}$ for the user to interact with $P_t$ under the assumption that the user values the product feature $A_{t-1}$, where 3 means the most likely and 0 means the least probable.
%
%
%
\noindent\textbf{\\[4px] \textsc{Task} 3: \taskIII}. To ensure that the proposed intent is reasonable and to guard against potential hallucinations, the third task asks the model to justify whether the proposed $C_t$ provides a reasonable explanation for the user to interact with $P_t$ after seeing $P_{t-1}$. Formally, the model is tasked to output a score $\mathcal{S}_3(C_t, P_{t-1}, I_{t-1}, P_t, I_t)\mid_{\mathcal{H}_{t-1}}$ $\in$ $\{0, 1, 2, 3\}$ indicating the plausibility of the generated comparison. 
%
%
%
\noindent\textbf{\\[4px] \textsc{Task} 4: \taskIV}. The final task we propose aims to test the model's ability to help recommendation systems decide whether to further recommend similar products or not. Providing the model with all the historical information and inferred purchasing intent, we ask it to choose from exposing the user to (a) similar products under the same category, (b) products with different features but still under the same category, (c) products under a different category (exploring further to infer user preferences). If we map the choices to the numerical scores $\{1,2,3\}$, then we formalize the task as questioning for $\mathcal{S}_4(exploration, I_{t})\mid_{\mathcal{H}_{t}}$ $\in$ $\{1, 2, 3\}$. Note that the degree of exploitation decreases and exploration increases as the score increases.

Tasks 1--3 are annotated on a four-point ordered scale, while Task 4 uses a three-way exploration choice. For evaluation, these raw annotations are later mapped to binary labels; we describe this mapping in Section~\ref{sec:eval-analysis} and Appendix~\ref{sec:task_design_explain}.

\subsection{Dataset}
We construct \dataset{} from Amazon-M2~\cite{Amazon-M2} and product images retrieved from the Amazon Review Dataset~\cite{Amazon-Review-Dataset}. Amazon-M2 provides session sequences and rich textual metadata such as product titles, prices, colors, and materials. We align those products with their corresponding images from the Amazon Review Dataset to obtain multimodal session records. After filtering out products with missing or inaccessible image links, we retain 10,905 sessions with complete textual and visual information.

\section{\dataset{} Construction}
We construct \dataset{} by enriching raw shopping sessions with explicit intention metadata. The pipeline in Figure~\ref{fig:maingraph} has four stages: (i) multimodal attribute extraction for each product, (ii) intention generation over the session timeline, (iii) metadata analysis of why intention shifts from one step to the next, and (iv) human annotation of a sampled subset for evaluation.

\subsection{Multi-modal Attribute Extraction}
The first stage extracts product attributes that can support later intention reasoning. We use GPT-4o-mini~\cite{GPT4omini} as the extraction model and provide both textual product descriptions and product images. The model outputs a coarse product category together with normalized attribute--value pairs, for example \textit{color: white} or \textit{size: 7.5 inches}. This step standardizes heterogeneous product metadata before any session-level reasoning is introduced.

\subsection{Customer Intention Generation}
To build up the intention tree based on the product purchase session, we first fill up the tree bones with predicted user intentions using L(V)LMs. The intentions are inferred at each time step following the session time frame. Starting with the first item in the session, we ask the model to infer a list of possible intentions $<I_{t1}, I_{t2}, I_{t3}, \ldots > \big|_{t=1}$ based on the textual and visual information of the product the user interacted with, where the prompt is demonstrated below. Then, we repeat the inference at every step as we add the next new session product into the visible list of items for the model.

To make the intention instantiation successional, we add the intention information of the previous time step $\{I_i \}_{i=1}^{t-1}$ (\textbf{\texttt{\textcolor{headcolor}{<Prev Intent>}}}) to facilitate the model's reasoning. At each time we perform the inference, we only use one intention chosen from the previous step's intentions to ensure coherent intention trajectory sampling.
More specifically, the model is constrained to output the five most possible user intentions, denoted as \textcolor{tailcolor}{\textbf{ \texttt{$\{$<New Intent i>$\}_{i=1}^{5}$}} }, prior to the fifth product at each iteration. This process is referred to as \textit{branching}, as it resembles the growth of a tree, wherein each new intention branches out from the initial concept, akin to twigs dividing into finer branches. Starting from the fifth product, we only infer one possible intention at a time to control the exponential growth of the tree size (by setting \textcolor{tailcolor}{\textbf{ \texttt{$\mid$<New Intent>$\mid$=1}} }).

\begin{center}
\vspace{-0in}
\resizebox{1\linewidth}{!}{
\begin{tabular}{l}
\textbf{\texttt{<TASK-PROMPT>}}\\
\textbf{\texttt{\textcolor{headcolor}{<INPUT:>}}}\\
\textbf{\texttt{\textcolor{headcolor}{<Prev Intent>}\textcolor{headcolor}{<Prev Products>}\textcolor{headcolor}{<New Product>}}}\\
\textbf{\texttt{\textcolor{tailcolor}{<OUTPUT:>}}}\\
\textbf{\texttt{\textcolor{tailcolor}{<New Intent 1>}\textcolor{tailcolor}{<Attr 1>}\textcolor{tailcolor}{<Rationale 1>}\textcolor{tailcolor}{<Comp 1>}}}\\
\textbf{\texttt{\textcolor{tailcolor}{<New Intent 2>}\textcolor{tailcolor}{<Attr 2>}\textcolor{tailcolor}{<Rationale 2>}\textcolor{tailcolor}{<Comp 2>}}}\\
~~\ldots~~\\
\textbf{\texttt{\textcolor{tailcolor}{<New Intent 5>}\textcolor{tailcolor}{<Attr 5>}\textcolor{tailcolor}{<Rationale 5>}\textcolor{tailcolor}{<Comp 5>}}}\\
\textbf{\texttt{\textcolor{headcolor}{<INPUT:>}}}\\
\textbf{\texttt{\textcolor{headcolor}{<Prev Intent>}\textcolor{headcolor}{<Prev Products>}\textcolor{headcolor}{<New Product>}}}\\
\textbf{\texttt{\textcolor{tailcolor}{<OUTPUT:>}}}\\
\end{tabular}
}
\end{center}

\begin{table}[t]
\small
\centering
\resizebox{!}{!}{
\begin{tabular}{@{}ll|ll@{}}
\toprule[1.5pt]
\textbf{Genre} & \textbf{Property} & \textbf{Train} & \textbf{Test} \\ 
\midrule[1.5pt]
\multirow{4}{*}{\textbf{Basic Info}} 
& \# Sessions (uni.) & 8963 & 5306 \\
& \# Sampled Tasks & 28736 & 7184 \\
& Avg. \# Products & 3.4163 & 3.4123 \\
& Avg. \# Intention & 3.4163 & 3.4123 \\
\midrule
\multirow{3}{*}{\textbf{Session Len}} 
& \# $Len=3$ & 18956 & 4752 \\
& \# $Len=4$ & 7598 & 1902 \\
& \# $Len=5$ & 2182 & 530 \\ 
\midrule
\multirow{4}{*}{\textbf{Task Num}}
& \# \textsc{Task 1} & 7153 & 1827 \\
& \# \textsc{Task 2} & 7171 & 1809 \\
& \# \textsc{Task 3} & 7154 & 1826 \\ 
& \# \textsc{Task 4} & 7258 & 1722 \\ 
\bottomrule[1.5pt]
\end{tabular}
}
\caption{Statistics of the sampled and human-annotated subset used for \dataset{}. \textit{uni.} denotes unique sessions. Sampling is performed over intention trajectories (and therefore task instances), not directly over raw sessions.}
\label{tab:benchmark_statistics}
\end{table}

\subsection{Intention-Shift Metadata Analysis}
Following this, we want to investigate the specific reasons behind each intention shift before and after the customer sees each product and how that might influence the customer's further decision-making. The prompt we used for generation is given above. To ground the reasoning in the actual product metadata, we require the model to point out the most likely feature \textbf{\texttt{\textcolor{tailcolor}{<Attr>}}} $A_t$ that affects the user's choices. 
Furthermore, we ask for a more comprehensive comparison \textbf{\texttt{\textcolor{tailcolor}{<Comp>}}} $C_t$ between the last product $P_t$ and the previous one $P_{t-1}$, so that it provides logical support for the modeled intention pathways.
To help models reason better, we require the model to provide rationales (\textbf{\texttt{\textcolor{tailcolor}{<Rationale>}}}) behind the generations as part of the output.
We collect this analysis metadata in the format of one general categorization plus one detailed instantiation, e.g., \textit{book type: fiction, price: \$20}.

\subsection{Human Annotation}
We hire Amazon Mechanical Turk annotators to label a randomly sampled subset of our data to balance cost and quality. We ask the workers to annotate with emphasis on the following perspectives: (1) the alignment of the proposed intention $I_t$ and session products $P_{t+1}$; (2) the consistency between the inferred valued attribute $A_t$ and the actual interacted products $P_{t+1}$; (3) the plausibility of the generated intention comparison $C_t$; (4) predictions on further intention pathways based on historical information.
In this way, the session intention could not only provide insights into the thinking process of customers but also meaningful references for when to explore and when to exploit product recommendation systems.
To simplify the annotation process, the annotators are only asked to assign a likelihood score or plausibility score for each task in a format roughly similar to \textit{yes, maybe yes, maybe no, no} (corresponding to $\mathcal{S} = 3, 2, 1, 0$).
We carried out multiple rounds of annotation worker selection with different criteria to ensure high annotation quality. We further analyze label distributions and worker behavior in Appendix~\ref{sec:append_annotation_detail}. Task 2 remains the most subjective task because it requires workers to judge whether a proposed valued attribute actually drives the next interaction, which is one reason we report detailed annotation analyses in the appendix.



\begin{table*}[t]
    \small
    \centering
    \resizebox{1\linewidth}{!}{
	\begin{tabular}{@{}l||cccccccc@{}}
	\toprule
    \multirow{2}{*}{\textbf{Models}}&\multicolumn{2}{c}{\textbf{Intent-Based Inference}} &\multicolumn{2}{c}{\textbf{Valued Attributes Reg.}}&\multicolumn{2}{c}{\textbf{Comparison Just.}}&\multicolumn{2}{c}{\textbf{Evolution Modeling}}\\
    \cmidrule(lr){2-3}\cmidrule(lr){4-5}\cmidrule(lr){6-7}\cmidrule(lr){8-9}
	&\textbf{Acc}&\textbf{Ma-F1}&\textbf{Acc}&\textbf{Ma-F1}&\textbf{Acc}&\textbf{Ma-F1}&\textbf{Acc}&\textbf{Ma-F1}\\
            \midrule
            \textsc{Random} & 50.00 & 50.00 & 50.00 & 50.00 & 50.00 & 50.00 & 54.38 & 35.00 \\
            \textsc{Majority} & 62.30 & 76.77 & 54.35 & NaN & 71.80 & 83.58 & 63.15 & NaN \\
            \bottomrule
            \rowcolor[gray]{0.9} \multicolumn{9}{c}{\textbf{LLM \textit{(Zero-Shot)}}}  \\
            \toprule
             Meta-Llama-3.1-8B & 56.87 & 70.98 & 49.36 & 55.10 & \underline{71.30} & \underline{83.24} & 39.26 & 53.01 \\
             Meta-Llama-3.2-3B & 54.68 & 63.97 & 52.02 & 43.48 & 33.13 & 49.48 & 51.34 & 36.61 \\
             Gemma-2-9B & 57.03 & 69.37 & 52.18 & 49.44 & 41.68 & 44.19 & \underline{53.77} & 34.54 \\
             Mistral-7B-v0.3 & \underline{\textbf{62.17}} & \underline{76.52} & 47.65 & \underline{\textbf{64.08}} & \underline{71.30} & \underline{83.24} & 39.61 & 53.53 \\
             Ministral-8B & 56.98 & 69.33 & 51.58 & 50.48 & 68.02 & 80.48 & 38.27 & 54.08 \\
             Mistral-Nemo-12B & 53.09 & 63.82 & 51.63 & 35.04 & 56.79 & 69.71 & 47.15 & 45.11 \\
             Falcon-3-7B & 57.31 & 71.74 & \underline{52.24} & 49.17 & 67.36 & 79.41 & 44.36 & 49.68 \\
             Falcon-3-10B & 54.95 & 66.93 & 51.35 & 48.59 & 65.49 & 78.24 & 43.84 & 45.89 \\
             Qwen-2.5-3B & 54.19 & 64.42 & 51.96 & 41.87 & 68.62 & 81.01 & 37.63 & \underline{53.98} \\
             Qwen-2.5-7B & 58.62 & 71.92 & 51.02 & 56.18 & 70.59 & 82.61 & 40.07 & 51.86 \\
            \bottomrule
            \rowcolor[gray]{0.9} \multicolumn{9}{c}{\textbf{LVLM \textit{(Zero-Shot)}}}  \\
            \toprule
             LLaVA-v1.6-mistral-7b & 58.29 & 71.90 & 47.48 & 62.27 & 62.94 & 75.11 & 37.62 & 54.20 \\
             LLaVA-v1.6-vicuna-7b & \underline{62.01} & \underline{76.55} & 46.93 & \underline{63.88} & \underline{71.27} & \underline{83.22} & 37.21 & \underline{\textbf{54.24}} \\
             Qwen-2-VL-7B & 58.73 & 71.48 & \underline{50.63} & 56.37 & 70.61 & 82.73 & \underline{37.67} & 53.95 \\
             Meta-Llama-3.2-11B-V & 45.10 & 61.38 & 38.41 & 52.35 & 42.11 & 59.20 & 36.33 & 53.23 \\
            \bottomrule
            \rowcolor[gray]{0.9} \multicolumn{9}{c}{\textbf{L(V)LM \textit{(Few-Shots)}}}  \\
            \toprule
             Mistral-7B-v0.3 & \underline{60.43} & 74.60 & \underline{50.64} & 61.39 & \underline{67.09} & 79.08 & \underline{43.44} & 49.85 \\
             Qwen-2-VL-2B & 58.02 & 73.46 & 40.63 & 58.40 & 66.70 & 79.92 & 36.99 & 53.45 \\
             LLaVA-v1.6-vicuna-7b & 51.06 & \underline{\textbf{77.26}} & 22.61 & \underline{62.92} & 66.81 & \underline{82.99} & 27.99 & \underline{54.07} \\
            \bottomrule
            \rowcolor[gray]{0.9} \multicolumn{9}{c}{\textbf{L(V)LM \textit{(Fine-tuned)}}}  \\
            \toprule
             Meta-Llama-3.1-8B & 52.82 & 63.84 & 51.46 & 46.27 & 70.76 & 82.82 & 51.92 & 33.01 \\
             Meta-Llama-3.2-3B & 55.67 & 66.80 & 51.80 & 46.70 & 69.61 & 81.93 & 51.63 & 32.66 \\
             Mistral-7B-v0.3 & 57.47 & 68.56 & 50.64 & 44.64 & 67.69 & 79.88 & 55.69 & 31.69 \\
             Ministral-8B & \underline{58.35} & \underline{69.55} & 51.24 & 45.01 & 66.54 & 79.10 & 55.57 & 35.11 \\
             Mistral-Nemo-12B & 56.10 & 66.80 & 52.02 & 46.68 & 67.74 & 79.81 & \underline{55.81} & 32.95 \\
             Qwen-2.5-7B & 54.02 & 65.63 & 52.02 & 46.75 & 69.50 & 81.66 & 54.47 & 31.59 \\
             Falcon-3-7B & 55.77 & 65.02 & \underline{\textbf{52.85}} & \underline{48.46} & \textbf{\underline{71.41}} & \underline{\textbf{83.30}} & 54.65 & \underline{36.86} \\
            \bottomrule
            \rowcolor[gray]{0.9} \multicolumn{9}{c}{\textbf{L(V)LM \textit{(Proprietary API)}}}  \\
            \toprule
            GPT4o-mini & 57.44 & 69.34 & 51.95 & 43.81 & \underline{71.19} & \underline{83.13} & 38.39 & \underline{53.90} \\
            GPT4o-mini (5-shots) & \underline{58.83} & \underline{71.86} & 49.32 & \underline{53.01} & 65.25 & 78.11 & 46.51 & 46.96 \\
            GPT4o-mini (COT) & 57.26 & 69.02 & 51.87 & 43.33 & 68.86 & 81.22 & 42.81 & 49.42 \\
            GPT4o & 55.05 & 65.33 & 49.75 & 36.27 & 56.30 & 67.51 & 41.64 & 52.39 \\
            GPT4o (5-shots) & 53.10 & 63.58 & 44.20 & 38.61 & 54.94 & 65.01 & 43.44 & 48.41 \\
            GPT4o (COT) & 53.30 & 61.91 & \underline{52.00} & 36.08 & 49.50 & 50.87 & \underline{\textbf{58.42}} & 13.73 \\
		\bottomrule
	\end{tabular}
    }
\caption{Evaluation results (\%) of different L(V)LMs on the human-annotated \dataset{} test set. Random samples labels uniformly. Majority always predicts the most frequent label for each task. All ``Few-shots'' results use 5 demonstrations. Within each block, the best scores are \underline{underlined}; the best scores overall are \textbf{bold}.} 
\label{tab:maintable_eval_results}
\end{table*}

\begin{table*}[t]
\small
\centering
\resizebox{1\linewidth}{!}{
\begin{tabular}{@{}ll|cccccccc@{}}
\toprule[1.5pt]
    \multirow{2}{*}{\textbf{Training Data}}&\multirow{2}{*}{\textbf{Backbone}}&\multicolumn{2}{c}{\textbf{Intent-Based Inference}} &\multicolumn{2}{c}{\textbf{Valued Attributes Reg.}}&\multicolumn{2}{c}{\textbf{Comparison Just.}}&\multicolumn{2}{c}{\textbf{Evolution Modeling}}\\
    \cmidrule(lr){3-4}\cmidrule(lr){5-6}\cmidrule(lr){7-8}\cmidrule(lr){9-10}
    & &\textbf{Acc}&\textbf{Ma-F1}&\textbf{Acc}&\textbf{Ma-F1}&\textbf{Acc}&\textbf{Ma-F1}&\textbf{Acc}&\textbf{Ma-F1}\\
\midrule[1.5pt]
\multirow{6}{*}{\textbf{\begin{tabular}[c]{@{}l@{}}Zero-shot\end{tabular}}}
     & Llama-3.1-8B & 56.87 & 70.98 & 49.36 & 55.10 & \underline{71.30} & \underline{83.24} & 39.26 & 53.01 \\
     & Llama-3.2-3B & 54.68 & 63.97 & 52.02 & 43.48 & 33.13 & 49.48 & \underline{51.34} & 36.61 \\
     & Mistral-7B-v0.3 & \underline{\textbf{62.17}} & \underline{\textbf{76.52}} & 47.65 & \underline{64.08} & \underline{71.30} & \underline{83.24} & 39.61 & 53.53 \\
     & Ministral-8B & 56.98 & 69.33 & 51.58 & 50.48 & 68.02 & 80.48 & 38.27 & \underline{\textbf{54.08}} \\
     & Falcon-3-7B & 57.31 & 71.74 & \underline{52.24} & 49.17 & 67.36 & 79.41 & 44.36 & 49.68 \\
     & Qwen-2.5-7B & 58.62 & 71.92 & 51.02 & 56.18 & 70.59 & 82.61 & 40.07 & 51.86 \\
\midrule
\multirow{6}{*}{\textbf{\begin{tabular}[c]{@{}l@{}}SIB\end{tabular}}}
     & Llama-3.1-8B & 52.82 & 63.84 & 51.46 & 46.27 & 70.76 & 82.82 & 51.92 & 33.01 \\
     & Llama-3.2-3B & 55.67 & 66.80 & 51.80 & 46.70 & 69.61 & 81.93 & 51.63 & 32.66 \\
     & Mistral-7B-v0.3 & 57.47 & 68.56 & 50.64 & 44.64 & 67.69 & 79.88 & \underline{55.69} & 31.69 \\
     & Ministral-8B & \underline{58.35} & \underline{69.55} & 51.24 & 45.01 & 66.54 & 79.10 & 55.57 & 35.11 \\
     & Qwen-2.5-7B & 54.02 & 65.63 & 52.02 & 46.75 & 69.50 & 81.66 & 54.47 & 31.59 \\
     & Falcon-3-7B & 55.77 & 65.02 & \underline{52.85} & \underline{48.46} & \underline{\textbf{71.41}} & \underline{\textbf{83.30}} & 54.65 & \underline{36.86} \\
\midrule
\multirow{6}{*}{\textbf{\begin{tabular}[c]{@{}l@{}}MIND + SIB\end{tabular}}}
     & Llama-3.1-8B & \underline{60.10} & 68.81 & 55.33 & 48.67 & 70.54 & 82.54 & 57.72 & 39.74 \\
     & Llama-3.2-3B & 59.88 & 67.92 & 55.28 & 50.15 & 64.02 & 75.48 & 58.54 & 40.50 \\
     & Mistral-7B-v0.3 & 60.04 & \underline{69.96} & 52.90 & 45.87 & 67.69 & 79.56 & \underline{\textbf{59.93}} & 37.16 \\
     & Ministral-8B & 58.24 & 67.33 & 53.95 & 47.44 & 65.44 & 77.01 & 58.77 & \underline{40.93} \\
     & Qwen-2.5-7B & 59.00 & 67.65 & 53.95 & 48.62 & 63.09 & 74.98 & 57.84 & 39.30 \\
     & Falcon-3-7B & 58.57 & 68.42 & \underline{\textbf{55.94}} & \underline{\textbf{50.22}} & \underline{71.30} & \underline{83.25} & 58.36 & 40.00 \\
\bottomrule[1.5pt]
\end{tabular}
}
\caption{Sequential finetuning with MIND followed by \dataset{}. ``MIND + SIB'' means that a backbone is first fine-tuned on MIND and then fine-tuned on \dataset{} (abbreviated as SIB). Within each training setting, the best scores are \underline{underlined}; the best scores overall are \textbf{bold}.}
\label{tab:transfer_MIND}
\end{table*}
\section{Evaluations and Analyses}
\label{sec:eval-analysis}
\subsection{Intrinsic Evaluations}
We present our detailed statistics in Table~\ref{tab:benchmark_statistics}.
By filling up the tree with intentions across 10,905 sessions, we obtain more than 1,950,000 intention entries and 1,100,000 intention trajectories. 
The majority of these sessions contain fewer than four products, though long sessions also exist with up to 18 products.
To sample a subset of sessions to form the \dataset{}, we first retrieve candidate sessions with lengths of three to five. 
We then sample 2,000 sessions with 2 trajectories per session and later add another disjoint 1,445 sessions with 4 trajectories per session. This gives 9,780 trajectories in total. To grant the model full information availability, we only query the tasks at the end of each session time step, that is, using all the available products and masking the last product when querying \textsc{Task} 1 and 2.

\subsection{Baselines and Model Selections}
\noindent\textbf{Evaluation protocol.} We report accuracy and Macro-F1. Although the raw annotations are ordinal, we evaluate in a binary setting to reduce neutral-response bias and to align the tasks with practical accept/reject-style decision making. For Tasks 1--3, answers $A/B$ are treated as positive and $C/D$ as negative. For Task 4, answer $A$ (continue exploiting similar products) is treated as positive and $B/C$ as negative. We include two simple baselines: \textsc{Random}, which samples labels uniformly, and \textsc{Majority}, which always predicts the globally most frequent label for a task. The Macro-F1 values for Majority on Tasks 2 and 4 are undefined because the majority label is negative, so the baseline never predicts a positive instance.

\noindent\textbf{Model families.} We evaluate four groups of models. \textbf{(i) Open zero-shot models:} open LLMs and LVLMs from the Llama~\cite{2024-llama3-models}, Gemma~\cite{2024-Gemma2}, Mistral~\cite{2023-mistral7b}, Falcon~\cite{2023-falcon}, Qwen~\cite{2025-Qwen2.5}, LLaVA~\cite{2023-llava}, and Qwen-VL~\cite{2024-Qwen2-VL} families. \textbf{(ii) Few-shot models:} selected open L(V)LMs evaluated with 5 in-context demonstrations. \textbf{(iii) Fine-tuned open models:} representative sub-11B backbones from different model families, fine-tuned on SIB with supervised fine-tuning (SFT) and LoRA using LLaMA-Factory. \textbf{(iv) Proprietary APIs:} GPT-4o and GPT-4o-mini~\cite{2024-GPT4, GPT4omini} under zero-shot, 5-shot, and Chain-of-Thought prompting~\cite{2023-COT}. The detailed split strategy, prompt construction, and fine-tuning setup are described in Appendix~\ref{sec:appendix}. Table~\ref{tab:transfer_MIND} additionally reports \textbf{MIND + SIB}, which means sequential fine-tuning on MIND first and then on SIB.

\subsection{Main Evaluation Results}
\noindent\textbf{\textsc{\taskIV{}} (\textsc{Task 4}) is the most challenging task.} Our experiments show that the average accuracy of the zero-shot models on \textsc{Task 4} is 42.34\%. Compared to the second hardest task (\textit{\taskII{}}), on which models scored 49.63\%, there is a large gap of 7.29\% on \textsc{Task 4}. After being fine-tuned, all open models are able to achieve a minimum accuracy of 51.92\%, while the top-performing one (Mistral-Nemo-12B) scores 55.81\%, just above the \textsc{Random} vote accuracy. It is worth noting that GPT-4o with Chain-of-Thought prompting is able to achieve the highest rate of 58.42\% among all models and methods. This might be because the larger model size and the technique of enabling reasoning at run time could help the model better mimic the thinking process of a real-life customer. This result shows that more work needs to be done to improve the model's capability to capture long-term user intention trends.

\noindent\textbf{Fine-tuning can greatly improve poorly performing models, but struggles to help mediocre ones.} Poorly performing models, which we refer to as those that receive a low score compared to models under the same category in some evaluation tasks, can quickly acquire relevant capabilities by being fine-tuned on the training set before testing. For example, LLAMA-3.2-3B shows poor performance on \textsc{Task 3} (\textit{\taskIII{}}), but after being fine-tuned on \dataset{}, it shows a performance increase of 36.5\% and demonstrates outcomes comparable with other larger 7B or 8B models. Mediocre performing models, which we refer to as those that score near the highest among the models but still struggle to surpass the top accuracy records, benefit less from fine-tuning. Among the proposed tasks, the largest maximum accuracy increase from zero-shot to fine-tuned occurs in \textsc{Task 4}, with a lift of 2.04\% in the highest score. As a result of these two factors, the variance between different models shrinks after fine-tuning. See Appendix \ref{sec:finetuning_additional_discussion} for additional discussions on fine-tuning.

\noindent\textbf{LVLMs struggle to make good use of visual signals.} In comparison to LLMs, which only use textual signals as input, LVLMs can refer to image information to facilitate their question answering and inference reasoning. However, as shown in Table \ref{tab:maintable_eval_results}, the highest accuracy scores of LVLMs still lag behind those of LLMs. When evaluated on \textsc{Task 4} using direct zero-shot prompting, the best LVLM outcome is even behind the best LLM by a large gap of 11.27\%. Possible reasons include the low signal-to-noise ratio of the collected images, and the fact that sellers usually include more comprehensive and concise product features in text format.

\begin{figure}[t!]
    \centering
    \includegraphics[width=\linewidth]{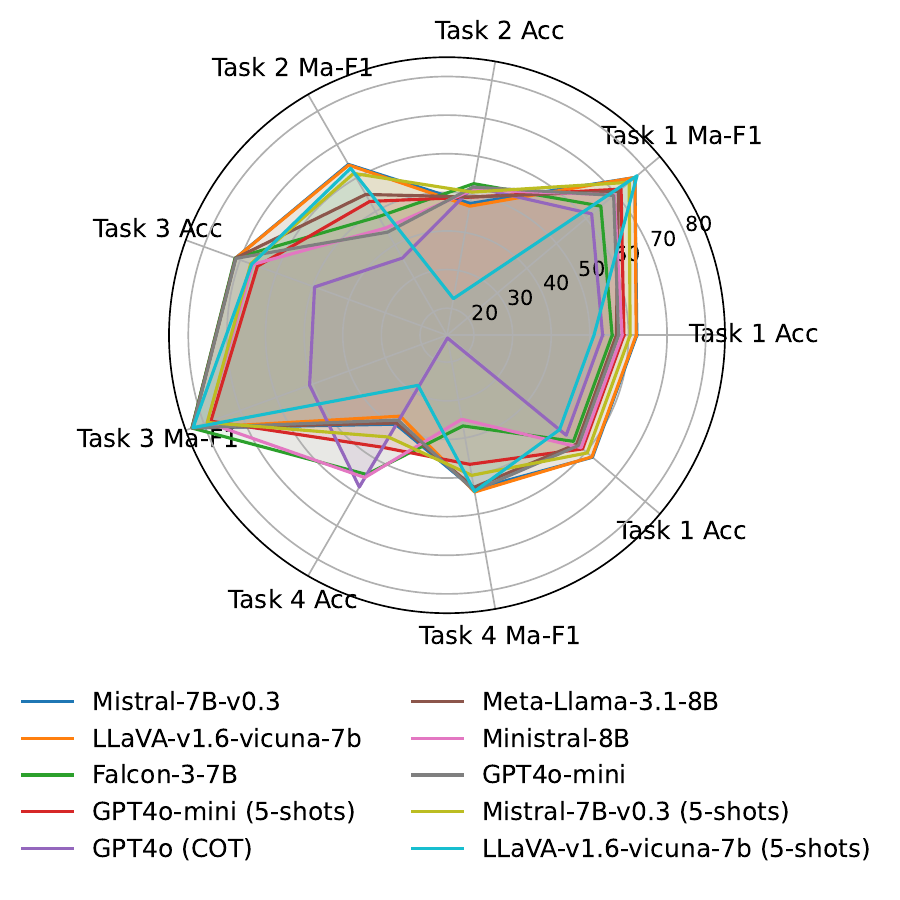}
    \caption{Radar chart for representative best-performing models from different evaluation settings. No single model dominates all four tasks.}
    \label{figs:model_radar_chart}
\end{figure}

\noindent\textbf{No model dominates across all tasks.} Figure~\ref{figs:model_radar_chart} visualizes representative top models from different settings. Mistral-7B-v0.3 is strongest among open zero-shot LLMs, LLaVA-v1.6-vicuna-7b is competitive among zero-shot LVLMs, and Falcon-3-7B performs best overall after SIB fine-tuning on several tasks. Yet none of them is uniformly best. This reinforces that \dataset{} probes multiple distinct capabilities rather than a single dominant skill.

\begin{figure}[t!]
    \centering
    \includegraphics[width=\linewidth]{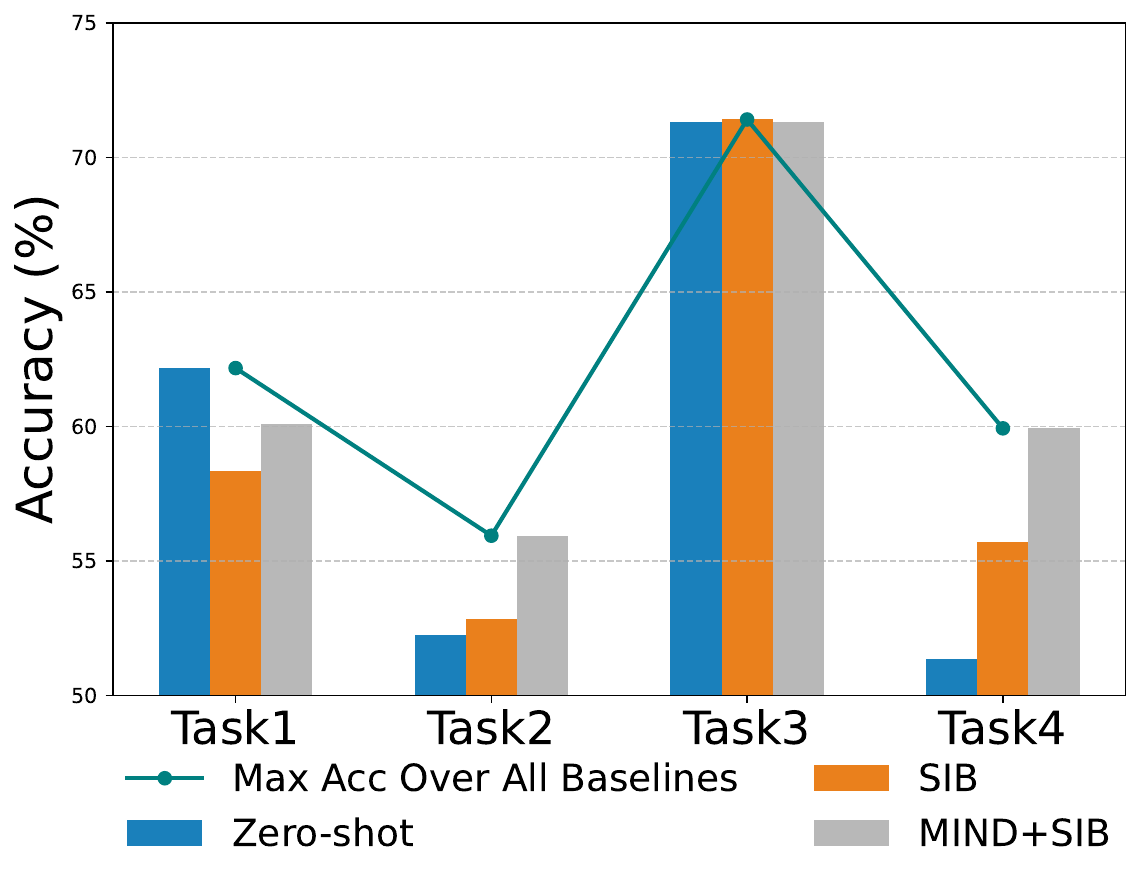}
    \caption{Comparison of the best performance across methods and tasks. The maximum accuracy achieved by the baselines is consistent with the maximum accuracy observed among open models across all methods (i.e., zero-shot, fine-tuning with \dataset{} (SIB), and sequential fine-tuning with MIND followed by SIB). In this setting, using proprietary APIs does not provide additional performance gains.}
    \label{figs:methods_comparison}
\end{figure}

\subsection{The Impact of Intention Injection}
\label{maintext:intention-injection}
From Table~\ref{tab:maintable_eval_results}, we observe that L(V)LMs struggle to directly leverage intention for next-product inference (\textit{Intent-Based Inference}) and to capture long-term shifts in intention from session history (\textit{Intention Evolution Modeling}). Table~\ref{tab:transfer_MIND} further examines whether generic intention knowledge transfers to, or assists with, answering questions in \dataset{}. We use MIND~\cite{MIND}, a co-buy intention resource, as an external source of intention supervision and sequentially fine-tune models on MIND and then SIB. This \emph{MIND + SIB} setting improves the best scores on Task~1 by 1.75 points, Task~2 by 3.09 points, and Task~4 by 4.24 points over SIB-only fine-tuning.

The gains are not uniform across tasks. Tasks~1, 2, and 4 depend directly on modeling latent intent and therefore benefit the most from intention injection. By contrast, Task~3 is already relatively strong in zero-shot and focuses more on local comparison consistency between adjacent products; it therefore has less room for improvement. We also evaluated models fine-tuned \emph{only} on MIND. Those models achieved less than 10\% accuracy on SIB because they did not reliably follow SIB's option-selection format, so we do not include them in the main table; details and examples are given in Appendix~\ref{appendix:further_notes_on_performances}.

\section{Model Performance Insights}
\label{further_notes_on_performances}
\subsection{Evaluation Task Performance Metrics}
We display the confusion-matrix statistics for GPT-4o with Chain-of-Thought prompting in Table~\ref{tab:task_performance_metrics}. Task~4 has the smallest true-positive count and the largest true-negative share, which is consistent with its overall difficulty: models often default to broader exploration judgments rather than correctly identifying when a session should remain in exploitation mode.
\begin{table}[ht]
\centering
\resizebox{1\linewidth}{!}{
\begin{tabular}{@{}llcccc@{}}
\toprule[1.5pt]
& \multirow{2}{*}{\textbf{Metric}} & \multicolumn{4}{c}{\textbf{Task No.}} \\ 
\cmidrule(l){3-6}
 & & \textsc{Task 1} & \textsc{Task 2} & \textsc{Task 3} & \textsc{Task 4} \\
\midrule[1.5pt]
\multirow{4}{*}{\textbf{Count}} 
&\# TP   & 781 & 415 & 527 & 57 \\
&\# FN   & 354 & 434 & 775 & 585 \\
&\# TN   & 234 & 515 & 309 & 949 \\
&\# FP   & 458 & 445 & 215 & 131 \\
\midrule
\multirow{4}{*}{\textbf{Percentage}} 
&TP (\%)  & 42.75\% & 22.94\% & 28.86\% & 3.31\% \\
&FN (\%)  & 19.38\% & 24.00\% & 42.44\% & 33.97\% \\
&TN (\%)  & 12.81\% & 28.47\% & 16.92\% & 55.11\% \\
&FP (\%)  & 25.07\% & 24.60\% & 11.77\% & 7.61\% \\
\bottomrule[1.5pt]
\end{tabular}
}
\caption{Task performance metrics for error analyses of GPT-4o with Chain-of-Thought on \dataset{}. TP, FN, TN, and FP denote true positive, false negative, true negative, and false positive predictions, respectively.}
\label{tab:task_performance_metrics}
\end{table}

\subsection{Finetuning}
\label{sec:finetuning_additional_discussion}
SIB-only fine-tuning is helpful for learning task format, but it does not always improve generalization. One reason is that SIB contains a broad and heterogeneous distribution of product categories, attributes, and intention trajectories. This makes the train--test gap relatively large even when the question format is fixed~\cite{DBLP:conf/acl/WangFXBSC23,DBLP:conf/emnlp/WangF0XLSB23,DBLP:conf/acl/0001FLS0XWBLJCS24,DBLP:conf/acl/0001S25,DBLP:conf/emnlp/WangFSXDZFBLLS25}. External intention supervision partially mitigates this issue: for example, Llama-3.1-8B drops from 56.87\% on Task~1 in zero-shot to 52.82\% after SIB-only fine-tuning, but improves to 60.10\% under sequential MIND + SIB fine-tuning.

\subsection{Error Analyses}
\label{maintext:error_analysis}
We randomly sample 200 error cases from GPT-4o with Chain-of-Thought prompting and ask three NLP PhD researchers to analyze them. The most common failure mode (47.5\%) is incorrect use of the provided metadata, especially failure to integrate earlier session context. Another 24\% of the sampled errors arise from annotation-task mismatches. We also observe failures to capture decisive product features (7\%), irrelevant or hallucinatory reasoning (6.5\%), and broader difficulty inferring the session's overall intent when the provided metadata is vague or weakly decisive (15\%). These percentages are computed within the error-analysis subset only; therefore, they should not be interpreted as estimates of the benchmark's overall noise rate.

\section{Conclusions}
In conclusion, we propose an automated pipeline to construct a large-scale knowledge base and further construct a sample dataset \dataset{} for L(V)LM evaluations. Extensive experiments show that current models struggle to understand and infer customers' intentions, while injecting intention information from other knowledge bases can improve performance. We hope our work can bridge the gap between intention understanding in simplified research cases like co-buy intention and more complex yet practical scenarios like session history. We also hope this framework can benefit the community by enabling better services with future models.

\section*{Limitations}
Our benchmark inherits limitations from both the source data and the curation process. First, a product interaction inside a session is only an imperfect proxy for user intention: users may click out of curiosity, because of presentation bias, or for reasons that are not reflected in the available metadata. Second, our intention tree is generated with GPT-4o-mini, so the benchmark may reflect generator bias even though we validate a subset through human annotation. Third, our current formulation does not use personalized signals such as long-term purchase history, demographics, or social context. Finally, Task~2 is inherently more subjective than the other tasks because which valued attribute best explains the observed transition is not always uniquely determined.

\section*{Ethics Statement}
\paragraph{Offensive Content Inspection}
We use publicly available e-commerce resources and model-generated metadata to build the benchmark. The generated metadata are grounded in product information and constrained by the session context. We do not ask models to produce free-form harmful content; the downstream evaluation tasks are structured classification problems.

\paragraph{Annotation Wage}
Annotators were recruited through Amazon Mechanical Turk. Workers participated voluntarily and were paid at an average hourly rate of approximately USD 15, in accordance with local requirements.

\paragraph{Licenses}
The Amazon-M2 dataset is released under the Apache 2.0 license. This grants us free access to the dataset.
Our code and data will be shared under the MIT license. It allows the free distribution of the assets we propose and curate. All associated licenses permit user access for research purposes, and we agree to follow all terms of use.

\section*{Acknowledgments}
The authors of this paper were supported by the ITSP Platform Research Project (ITS/189/23FP) from the Innovation and Technology Commission of Hong Kong SAR, China, and by the AoE (AoE/E-601/24-N), RIF (R6021-20), and GRF (16205322) from the Research Grants Council of Hong Kong SAR, China.
We also thank the Amazon Stores Foundational AI team for their support and for providing valuable insights on data curation and evaluation.

\bibliography{custom}

\newpage
\appendix
\begin{center}
    {\Large\textbf{Appendices}}
\end{center}

\section{Implementation Details}
\label{sec:appendix}
\label{sec:append_implement_detail}

\subsection{Attribute Extraction}
To extract product attributes with GPT-4o-mini, we use the following 3-shot prompt template:

\begin{displayquote}
\textbf{\texttt{\small 
Your goal is to extract the attribute type and attribute values of the product. \\
You will be provided with the product names and their corresponding product images, and you will output for the product: \\
Category: general category name of the product. Keep the category name simple and within 3 words. \\
Attributes: attribute(s) of the product. You can infer new ones from the image. Keep the attribute simple and within 3 words each. Separate different attributes by |. Generate in the format of attribute: value \\[8px] 
Below are three examples:
}}\\
\ldots \\
\textbf{\textcolor{headcolor}{\texttt{\small Input: \\
Product Name: Adidas Ultraboost 21 Women's Running Shoes on sale, White/Pink special, Size 8 only, best for daily runs!
}}} \\
\textcolor{tailcolor}{\texttt{\small\textbf{Output: \\
Category: Clothing \\
Attributes: brand: Adidas | model: Ultraboost 21 | gender: Women's | type: Running Shoes | color: White/Pink | size: 8
}}}\\
\textbf{\textcolor{headcolor}{\texttt{\small Input: \\
Product Name: Lightweight and powerful Dell XPS 13 Laptop, with newly released Intel i7, 16GB RAM, enhanced 512GB SSD, Silver version
}}} \\
\textcolor{tailcolor}{\texttt{\small\textbf{Output: \\
Category: Electronics \\
Attributes: brand: Dell | model: XPS 13 | processor: Intel i7 | RAM: 16GB | storage: 512GB | color: Silver
}}}\\
\textbf{\textcolor{headcolor}{\texttt{\small Input: \\
Product Name: baking enthusiasts' good friend - KitchenAid Artisan Series 5-Quart Stand Mixer, Empire Red
}}} \\
\textcolor{tailcolor}{\texttt{\small\textbf{Output: \\
Category: Kitchen Appliance \\
Attributes: brand: KitchenAid | model: Artisan Series | capacity: 5-Quart | type: Stand Mixer | color: Empire Red
}}}\\
\ldots \\
\textbf{\texttt{\small 
Input: \\
<INPUT MESSAGE> \\
Output: \\ }}
\end{displayquote}

\subsection{Intention Tree Construction}
To construct the intention tree and populate it with intentions, valued attributes, and supporting comparisons, we use the following 5-shot template:
\begin{displayquote}
\textbf{\texttt{\small 
Act as a customer who is browsing a series of products. \\
For each input, you are required to generate several intentions as output, and each intention should only contain the following lines of information: \\
New Intention: new intention you may have after interacting with the new product \\[2px]
Attribute: attribute(s) of the new product that caused the change in intention. You can infer new ones from the image. Generate in the format of attribute: value \\[2px]
Rationale: a short rationale explaining why the attribute of the new product reflects the new intention. Generate in the format of facets: reasoning \\[2px]
Comparison: a comparison between the new product and the previous product to justify why the new product caused the change in intention. Generate in the format of aspects: comparison \\[5px]
Here is one example with five intentions:
}}\\
\ldots \\
\textbf{\textcolor{headcolor}{\texttt{\small Input: \\
Previous Intention: Looking for stylish and modern footwear that complements their athletic look. \\
Previous Product: Nike Free Metcon 5 Women's Workout Shoes (varieties: runner, target consumers: women, size: 3.5, price: \$100). \\
New Product: LV Glove Loafer (varieties: loafer, target consumers: men, size: 3.5, price: \$200, structure: cushioned insole).
}}} \\
\textcolor{tailcolor}{\texttt{\small\textbf{Output: \\
New Intention: Invest in premium quality footwear for long-lasting style and comfort. \\
Attribute: design: luxury material and craftsmanship. \\
Rationale: durability: The LV Glove Loafer is crafted from high-quality materials, offering durability and style that ensures it will last longer than ordinary shoes. \\
Comparison: collectability: compared to the Nike Free Metcon 5, which focuses on performance, the LV offers a blend of luxury and longevity, making it a worthy investment. \\[3px]
New Intention: Own a versatile pair of shoes suitable for both casual and formal settings. \\
Attribute: varieties: loafer. \\
Rationale: usages: The loafer style of the LV Glove Loafer makes it versatile enough to be worn in both casual and formal settings, unlike the more specialized athletic design of the Nike Free Metcon 5. \\
Comparison: versatility: While the Nike Free Metcon 5 is primarily designed for workouts, the LV Glove Loafer's loafer style offers versatility for various occasions. \\[3px]
New Intention: Enhance your wardrobe with a statement piece that reflects personal style. \\
Attribute: design: unique and luxurious. \\
Rationale: aesthetics: The unique and luxurious design of the LV Glove Loafer makes it a statement piece that can elevate any outfit, reflecting personal style. \\
Comparison: uniqueness: Unlike the more common athletic design of the Nike Free Metcon 5, the LV Glove Loafer stands out as a unique and stylish addition to the wardrobe. \\[3px]
New Intention: Prioritize comfort without compromising on style. \\
Attribute: comfort: cushioned insole. \\
Rationale: comfort: The cushioned insole of the LV Glove Loafer ensures comfort for long periods, making it a practical choice without compromising on style. \\
Comparison: comfort: While the Nike Free Metcon 5 is designed for athletic performance, the LV Glove Loafer offers a balance of comfort and style for everyday wear. \\[3px]
New Intention: Choose a high-end brand to reflect social status. \\
Attribute: brand: Louis Vuitton. \\
Rationale: status: Owning a product from a high-end brand like Louis Vuitton reflects social status and prestige. \\
Comparison: brand prestige: Compared to Nike, which is known for athletic wear, Louis Vuitton is a luxury brand that signifies higher social status.
}}}\\
\ldots \\
\textbf{\texttt{\small 
Input: \\
Previous Intention: <Previous Intention>\\
Previous Product: <PREVIOUS PRODUCTS>\\
New Product: <THE LAST PRODUCT> \\
Output: \\}}
\end{displayquote}

For smaller branching factors, we simply reduce the number of requested outputs. For larger branching factors, we extend the prompt with additional examples as needed.

\subsection{Intention Generator Model Selection}
We first tested open LVLMs such as Mantis and LLaVA for metadata generation, but found them unreliable for large-scale intention-tree construction. The main issue was not raw semantic quality, but format control: many open models failed to follow the required output schema consistently when given long textual metadata together with images, occasionally degenerating into repeated tokens or malformed structures. GPT-4o-mini was substantially more stable while remaining cost-effective, so we use it as the primary generator for the intention tree.

\subsection{Fine-tuning Model Selection}
We select fine-tuning backbones using three criteria. \textit{(1) Organizational diversity:} we choose models from different model families to avoid overfitting conclusions to a single ecosystem. \textit{(2) Size constraint:} models must be smaller than 11B parameters so they can be fine-tuned on our hardware. \textit{(3) Representative strength:} within each family, we choose a strong and practically relevant checkpoint. For example, we use Llama-3.1-8B as Meta's representative 7B/8B backbone because later Llama-3.2 releases focus on smaller models and Llama-3.3 targets much larger ones.

\subsection{Training-Test Splits}
The detailed process is outlined as follows:
\textit{(1) Indexing:} We created an index for all annotated questions. Each questionnaire contained four questions corresponding to Tasks 1–4, ensuring an equal number of samples per task. Therefore, the proportions of indices for each question are equal.
\textit{(2) Index Set Creation:} A unified set of indices was constructed, where each index uniquely corresponds to a specific Task and Session number for traceability.
\textit{(3) Splitting:} We adopted a 4:1 train-test split. Indices were randomly sampled to create the training and test sets. Although the number of samples for Tasks 1--4 may vary slightly due to random sampling, the distribution remains largely balanced.
\textit{(4)} The resulting training and test sets were used across different models and training schemes (e.g., zero-shot, fine-tuning with SIB, and sequential fine-tuning with MIND followed by SIB)

\subsection{Few-shot Example Curation}
When curating few-shot demonstrations, we prioritize clarity over distributional coverage. Each demonstration should be concise, internally consistent, and easy for a human annotator to judge. We first generate a larger candidate pool with GPT-4o across several product categories, then manually filter, revise, and validate the final examples. We also test prompt variants with different example categories and example counts; on GPT-4o, these changes lead to only minor accuracy fluctuations (within 1\%), so we use a single 5-shot configuration throughout the paper.

\subsection{Model Evaluation}
We evaluate models with zero-shot prompts (Table~\ref{tab:appendix_eval_zeroshot_prompts}), 5-shot prompts (Tables~\ref{tab:appendix_eval_5shots_prompts_task1}--\ref{tab:appendix_eval_5shots_prompts_task4}), and Chain-of-Thought prompts (Table~\ref{tab:appendix_eval_CoT_prompts}). For consistency, every result described as ``Few-shots'' in Table~\ref{tab:maintable_eval_results} uses exactly five demonstrations.

\subsection{Finetuning Methods}
All fine-tuning experiments use supervised fine-tuning (SFT) with LoRA. We implement them with the open-source LLaMA-Factory framework~\cite{llamafactory}. 

\subsection{Role of GPT-4o-mini}
GPT-4o-mini is used in both dataset construction and proprietary-model evaluation. We include it in evaluation because it is the generator used to produce the intention tree, and we want to test whether generation-time familiarity translates into an unfair advantage. It does not: GPT-4o-mini is competitive, but not dominant, on the benchmark. To avoid confounding the error analysis with the generation model, we analyze GPT-4o Chain-of-Thought errors rather than GPT-4o-mini errors in Section~\ref{maintext:error_analysis}.

\subsection{Expert Selection for Error Analysis}
The error-analysis experts are three Computer Science PhD researchers from our institution, each with multiple publications in NLP or closely related areas. We randomly sample 200 incorrect GPT-4o-CoT predictions across Tasks~1--4. Each expert independently reviews the sampled cases using a shared error taxonomy, and the final labels are determined by consensus discussion.

\section{Theoretical Framework}
\label{sec:theory_and_explain}

\subsection{Intention Tree}
The intention tree $\mathbf{T}$ is defined inductively over the session timeline. At each time step $t \in \mathbb{N}^{+}$, we extend $\mathbf{T}_{1,\ldots,t-1}$ to $\mathbf{T}_{1,\ldots,t}$ by attaching one or more candidate intentions for the newly observed product. The information accumulated up to time step $t$ is denoted by $\mathcal{H}_t$.

Traditional formulations often predict the next interaction directly from the most recent product or intention state, for example through $\mathbb{P}(P_{t+1}\mid P_t)$ or $\mathbb{P}(I_{t+1}\mid I_t)$. Our formulation instead factorizes the problem through an intermediate session-level latent state:
\begin{equation}
\begin{aligned}
\mathbb{P}(P_{t+1}\mid \mathcal{H}_t)
=\, & \mathbb{P}(\mathcal{M}_t,P_{t+1}\mid \mathcal{H}_t) \\
=\, & \mathbb{P}(\mathcal{M}_t\mid \mathcal{H}_t)\cdot \mathbb{P}(P_{t+1}\mid \mathcal{H}_t,\mathcal{M}_t) \\
\approx\, & \mathbb{P}(\mathcal{M}_t^{\phi}\mid \mathcal{H}_t)\cdot \mathbb{P}(P_{t+1}\mid \mathcal{H}_t,\mathcal{M}_t^{\phi}),
\end{aligned}
\end{equation}
where $\mathcal{M}_t^{\phi}$ is the model's approximation of the latent session state $\mathcal{M}_t$.

Rather than representing this latent state as a monolithic variable, we decompose it into three explicit components:
\begin{itemize}
    \item $I_t^{\zeta}$: inferred intention,
    \item $A_t^{\zeta}$: valued attribute, and
    \item $C_t^{\zeta}$: comparison metadata.
\end{itemize}
Together they form
\[
\mathcal{M}_t^{\phi}=\left(A_t^{\zeta},I_t^{\zeta},C_t^{\zeta}\right),
\]
so the next-step reasoning problem becomes
\[
\mathbb{P}(P_{t+1}\mid \mathcal{H}_t,A_t^{\zeta},I_t^{\zeta},C_t^{\zeta}).
\]
The superscript $\zeta$ indicates that these components are branch-specific approximations generated by the intention-tree construction model.

This decomposition also motivates the four tasks. For example, in Valued Attribute Regularization we sample a subset of candidate attributes $\mathcal{A}^{\pi}\subseteq \mathcal{A}$ and ask the model to determine whether the transition to $P_{t+1}$ is plausible under the assumption that a valued attribute lies in $\mathcal{A}^{\pi}$. The task can therefore be seen as constraining the branch state so that $A_t^{\zeta}\in\mathcal{A}^{\pi}$ and then evaluating whether the resulting branch remains compatible with the observed session history.

\subsection{Intuition}
The branching process is easiest to understand with a short session. Suppose a session contains two products and uses a 5-branching scheme. We first infer one intention for the first product. After the second product is observed, the model proposes five plausible follow-up intentions. When a third product arrives, each of those intentions can in turn branch into additional candidate continuations. Every intention node is paired with one attribute, one rationale, and one comparison.

\section{Task and Evaluation Design}
\label{sec:task_design_explain}
\subsection{Design Criteria for Choice Options}
The 0--3 scores in the task definitions are symbolic encodings of concrete answer choices. For Tasks~1--3, score 3 corresponds to a strong positive judgment, score 2 to a weak positive judgment, score 1 to a weak negative judgment, and score 0 to a strong negative judgment. Table~\ref{tab:appendix_eval_zeroshot_prompts} provides the exact formulations used in prompting and annotation.

For evaluation, we merge adjacent options into binary labels. Specifically, for Tasks~1--3 we group $A/B$ together and $C/D$ together. This reduces neutral-response bias and makes the final prediction problem better aligned with practical decision settings. For Task~4, the positive class is option $A$ (continue recommending similar products), while $B/C$ are grouped as the negative class because they both indicate a need for broader exploration.

\subsection{Why Not Use Open-Ended Question Answering}
We considered an open-ended formulation but ultimately chose structured multiple-choice evaluation for three reasons. First, open-ended intention proposals are difficult to standardize and cluster at scale, which makes both benchmarking and human evaluation much less reliable. Second, open-ended annotation is substantially more expensive because it requires more expert labor per example. Third, our worker-selection process already shows that reliability is a challenge even for structured questions: among 300 initial candidates, only 11 passed the full quality-control pipeline. In our setting, a structured formulation offers the best trade-off between scalability, annotation quality, and reproducibility.

\section{Annotation Process}
\label{sec:append_annotation_detail}
\subsection{Worker Selection Protocol}
\label{appendix:worker_selection_protocol}
We apply a multi-stage quality-control pipeline to obtain reliable annotations. Qualification invitations are sent only to AMT workers with more than 2,000 approved HITs and an approval rate above 90\%. We then administer a qualification test built from sampled sessions with author-validated gold labels. Workers must score above 75\% while completing at least 20 questions to move forward.

After the qualification stage, we further remove workers who exhibit obvious low-effort behavior, especially those who choose the same side of the label space almost all the time. A second screening round reveals multiple such one-sided annotators, who are excluded before the main round. In the end, 11 workers remain out of 300 initial candidates, corresponding to a 3.67\% retention rate.

\subsection{Annotation Instructions}
\label{appendix:annotation_interface}
We present the annotation interface in non-technical language while keeping it closely aligned with the task definitions in Section~\ref{sec:task_definitions}. For the first three questions, workers assign a score on a four-point plausibility scale from 0 to 3. For the fourth question, workers choose among three exploration options, also mapped to an ordered scale. We explicitly explain the session-product list, the proposed intention metadata, and the meaning of each answer choice to reduce ambiguity.

\section{Annotation Result Analysis}

\subsection{Raw Label Result}
\label{appendix:raw_label_result}
The label distribution summarized in Table~\ref{tab:raw_label_result_distribution} corresponds to the Majority baseline in Table~\ref{tab:maintable_eval_results}. We report merged binary labels, for Tasks~1--3, $A/B$ and $C/D$ are grouped together, and for Task~4, $A$ is contrasted with $B/C$. We grouped them to mitigate individual annotator biases observed during the annotation process, where some annotators consistently favored extreme responses while others tended to choose intermediate options.

\begin{table}[ht]
\centering
\begin{tabular}{|c|c|r|c|}
\hline
\textbf{Task\_Ind} & \textbf{Label} & \textbf{Count} & \textbf{Percentage} \\
\hline
1 & A\_B & 5844 & 62.30\% \\
1 & C\_D & 3536 & 37.70\% \\
2 & A\_B & 4282 & 45.65\% \\
2 & C\_D & 5098 & 54.35\% \\
3 & A\_B & 6735 & 71.80\% \\
3 & C\_D & 2645 & 28.20\% \\
4 & A    & 3456 & 36.84\% \\
4 & B\_C & 5924 & 63.16\% \\
\hline
\end{tabular}
\caption{Merged label counts and percentages after binarization. \textit{Task\_Ind} denotes the task index. For Tasks~1--3, $A\_B$ denotes positive labels and $C\_D$ denotes negative labels; for Task~4, $A$ denotes the exploitative recommendation choice and $B\_C$ denotes the exploratory choices.}
\label{tab:raw_label_result_distribution}
\end{table}

\subsection{Consistency}
We establish the final ground truth by majority vote over three annotators. In Table~\ref{tab:Binary_Answer_consistency_board}, ``3:0'' means full agreement and ``2:1'' means one annotator disagrees with the other two after labels are mapped to the binary evaluation space. More than half of the questions receive full agreement in their binary labels, indicating that the benchmark is difficult but not annotation-random.

\begin{table}[ht]
\centering
\begin{tabular}{|c|r|r|r|}
\hline
\textbf{Task\_Ind} & \textbf{2:1} & \textbf{3:0} \\
\hline
1 & 6041  & 7959 \\
2 & 9170  & 4830 \\
3 & 5390  & 8610 \\
4 & 3934  & 10066 \\
\hline
\end{tabular}
\caption{Consistency analysis of binary answer label distribution. \textit{Task\_Ind} denotes the task index, ranging from 1 to 4.}
\label{tab:Binary_Answer_consistency_board}
\end{table}

\subsection{Annotation Quality Filter}
Beyond dataset-level statistics, we also inspect individual annotator behavior. Table~\ref{tab:One-side_Clicker_Board} illustrates a failure mode observed during worker screening: some workers overwhelmingly favor one side of the label space regardless of the example. We use this pattern as one of the exclusion criteria in the qualification pipeline, because it suggests low engagement rather than a principled annotation strategy.

\begin{table}[ht]
\centering
\begin{tabular}{|l|r|r|r|r|}
\hline
\textbf{Annotator\_ID} & \textbf{A} & \textbf{B} & \textbf{C} & \textbf{D} \\
\hline
A1***1A & 3201 & 719  & 893  & 31  \\
A2***EZ & 3402 & 1208 & 437  & 1   \\
A2***2M & 106  & 5540 & 1950 & 48  \\
A1***SU & 633  & 186  & 135  & 18  \\
A3***TX & 2113 & 173  & 610  & 28  \\
A2***BO & 287  & 32   & 49   & 0   \\
A2***YO & 919  & 221  & 196  & 24  \\
A2***E0 & 466  & 129  & 140  & 5   \\
AF***9P & 60   & 23   & 14   & 3   \\
\hline
\end{tabular}
\caption{Annotators excluded during screening because they overwhelmingly favored the same option across many questions.}
\label{tab:One-side_Clicker_Board}
\end{table}

\subsection{Benchmark and Data Quality Validation}
Some sessions are inherently ambiguous. Users may make abrupt jumps between products, and the generated intention metadata can occasionally expose those inconsistencies rather than resolve them. These cases are part of the difficulty of the benchmark, although additional preprocessing could reduce them further in future releases.

\subsection{Clarifications on the Majority Vote Score}
The Majority baseline in Table~\ref{tab:maintable_eval_results} is a task-level prediction baseline, \emph{not a measure of human performance}. For each task, we examine the final binary labels over the whole dataset and \emph{always predict the globally most frequent class}. For example, on Task~3 the majority class is the positive side ($A/B$), while on Tasks~2 and 4 it is the negative side. This baseline is useful because it reveals \emph{label skew}, but it should not be interpreted as evidence that a fixed answer is correct for every question.

\subsection{Missing F1 Score for Tasks 2 and 4}
The NaN Macro-F1 values for the Majority baseline on Tasks~2 and 4 are a consequence of the prediction distribution. Because the baseline always predicts the negative class on those tasks, it never produces a positive prediction; precision for the positive class is therefore undefined, which propagates to the F1 score. Accuracy remains well defined and is still reported.

\section{Model Performance Insights}
\label{appendix:further_notes_on_performances}
\subsection{Imbalanced Task Performance Gain with Intention Injection}
\label{appendix:intention-injection}
The gains from intention injection are largest on tasks that require the model to use latent intent directly. Task~1 asks whether a new product is compatible with a proposed intent, Task~2 asks whether a valued attribute explains the transition, and Task~4 asks whether the session has become exploratory. All three benefit from generic intention supervision. Task~3 is different: it primarily tests whether a generated comparison is locally coherent with adjacent products, so it already has a strong zero-shot signal and shows less room for improvement.

\subsection{Solely Fine-Tuning with MIND}
Models fine-tuned only on MIND perform poorly on SIB (below 10\% accuracy). The main issue is an output format mismatch. MIND is open-ended and encourages descriptive intent generation, whereas SIB requires selecting discrete options. As a result, MIND-only models tend to produce fluent but verbose reasoning without clearly stated answers,  leading to uniformly low performance across models (<10\%). This issue is further amplified by model capacity: small models used in our experiments (those under 11B parameters) have limited ability to rapidly adapt to a different answer format at inference time without additional task-specific fine-tuning. We intentionally keep MIND in the transfer study because its open-ended intent supervision is precisely what makes it a useful source of generic intention knowledge, but it must be followed by SIB fine-tuning to adapt to the benchmark format~\cite{wang2026arxiv2tablerealisticbenchmarkingevaluation,DBLP:journals/corr/abs-2601-22448}.

\begin{displayquote}
\textbf{\texttt{\small 
[{
    "instruction": "Act as a customer who bought these two product: ",
    "input": "Product A: Lincoln Stain Wax Shoe Polish 3 Fl Oz (Selection of Colors); Product B: Angelus Shoe Wax Polish 3fl Oz (Color Variety); What is your possible co-buy intention for these two products?",
    "output": "The potential co-buy intention could be that the person wants to purchase both shoe polish products to have a variety of colors to choose from when polishing their shoes"
},
{
    "instruction": "Act as a customer who bought these two product: ",
    "input": "Product A: BMC Mens 6 pc Mixed Design Self Tie Bowtie Pocket Square Suit Accessories; Product B: Tenby Living 2-Pack Black Tie Rack, Organizer, Hanger, Holder - Affordable Ti.; What is your possible co-buy intention for these two products?",
    "output": "The potential co-buy intention for people purchasing these two products simultaneously could be to enhance their wardrobe and maintain an organized and stylish appearance"
},
 ...]
}}
\end{displayquote}

\subsection{The BERT-based Models}
We also test strong pretrained encoder baselines such as RoBERTa-large-355M and DeBERTa-v3-large, but do not include them in the main comparison because they fail to follow the task format reliably.

For RoBERTa-large-355M, the raw output is exemplified as follows:

\begin{displayquote}
\textbf{\texttt{\small 
[{"task\_counter": 25248, "session\_counter": 6311, "question\_idx": 3, "response": "**A**"}, {"task\_counter": 27563, "session\_counter": 6890, "question\_idx": 2, "response": "**Yes**"}, {"task\_counter": 2654, "session\_counter": 663, "question\_idx": 1, "response": "**A**"}, {"task\_counter": 16969, "session\_counter": 4242, "question\_idx": 0, "response": "**A**"}, {"task\_counter": 33507, "session\_counter": 8376, "question\_idx": 2, "response": "**Yes**"}, ...]
}}
\end{displayquote}

RoBERTa often defaults to generic answers such as ``A'' or ``Yes'' regardless of the question, suggesting that it is not grounding its prediction in the full session context.

For DeBERTa-v3-large, the raw output often consists of malformed strings such as ``IBILITY'' and ``Measurement'' rather than valid options:

\begin{displayquote}
\textbf{\texttt{\small 
[{"task\_counter": 25248, "session\_counter": 6311, "question\_idx": 3, "response": "**IBILITY**"}, {"task\_counter": 27563, "session\_counter": 6890, "question\_idx": 2, "response": "**IBILITY**"}, {"task\_counter": 2654, "session\_counter": 663, "question\_idx": 1, "response": "**Measurement**"}, {"task\_counter": 16969, "session\_counter": 4242, "question\_idx": 0, "response": "**IBILITY**"}, {"task\_counter": 33507, "session\_counter": 8376, "question\_idx": 2, "response": "**IBILITY**"}, ...]
}}
\end{displayquote}

These failures suggest that standard encoder-only models are not well suited to the instruction-following setting required by \dataset{}.

\begin{figure*}[t]
    \centering
    \includegraphics[width=0.9\linewidth]{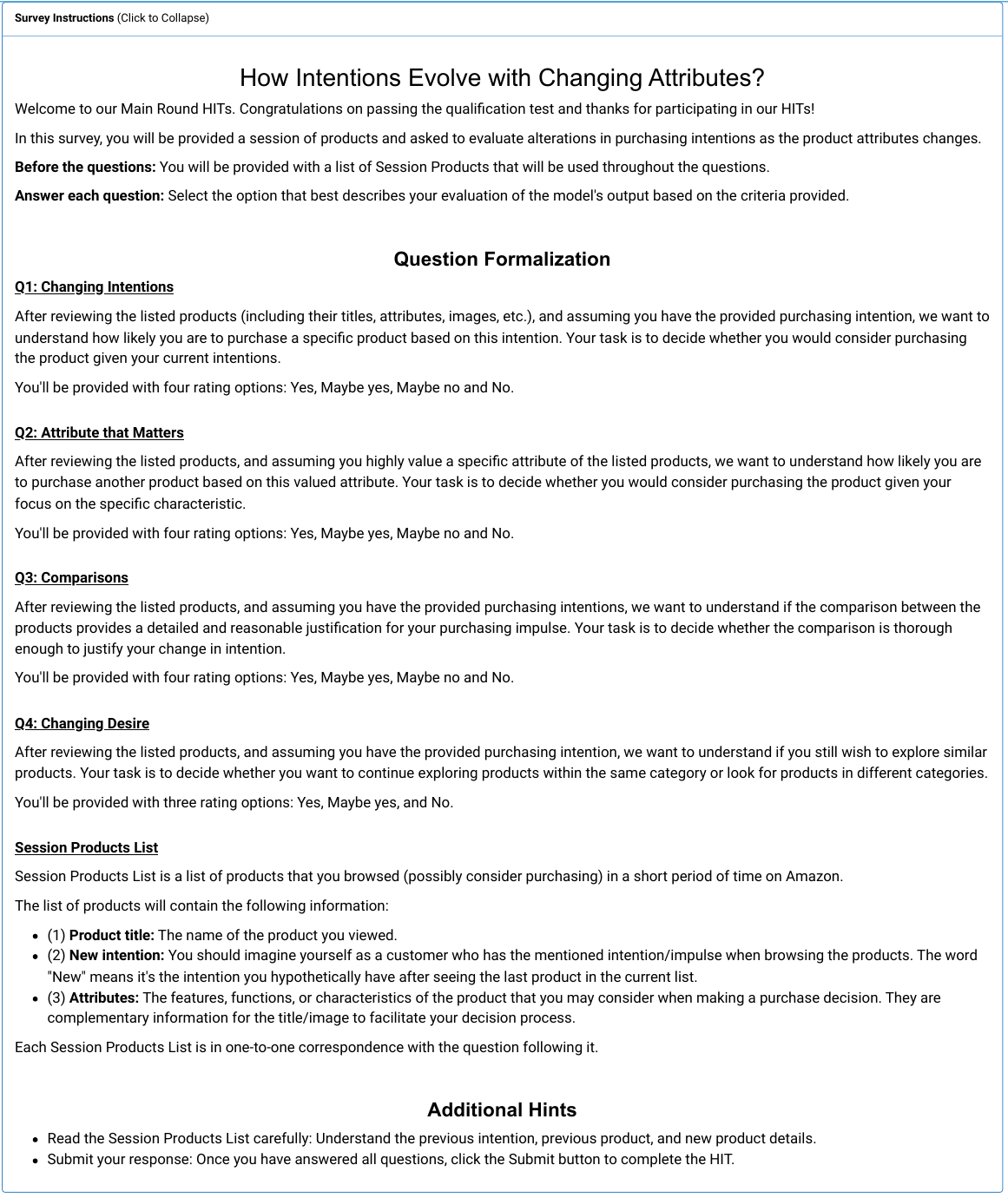}
    \vspace{-0in}
    \caption{Annotation instructions shown to workers. The interface explains the task definitions in plain language and previews the information available in the session-product list.}
    \label{fig:survey_instruction}
    \vspace{-0.2in}
\end{figure*}



\begin{table*}[t]
\small
\centering
\resizebox{\linewidth}{!}{
\begin{tabular}{@{}l|l@{}}
\toprule
Task & Zero-shot Prompt \\ 
\midrule
\textsc{Task 1} & \begin{tabular}[c]{@{}l@{}}
Act as a customer who is browsing a series of products given as follows. \\
\texttt{<session product information>} \\[7px]

After seeing \texttt{<previous products>}, \\
and assuming you are a customer who has the intention of \texttt{<second last intention>}. \\
How likely are you to purchase \texttt{<last product>} based on the assumed intention? \\[7px]

A. Yes: The product is a logical and reasonable outcome of the purchasing intention. \\
B. Maybe yes: I may consider this, but it's not a strong impulse. \\
C. Maybe no: The product is not directly related to my intention. \\
D. No: I would never purchase it if I were the customer with the given intention. \\[7px]

Your Answer (Answer A or B or C or D only): \\
\end{tabular}\\ 
\midrule
\textsc{Task 2} & \begin{tabular}[c]{@{}l@{}}
Act as a customer who is browsing a series of products given as follows. \\
\texttt{<session product information>} \\[7px]

After seeing \texttt{<previous products>}, \\
and assuming you are a customer who highly value the feature \texttt{<second last intention attribute>} \\
of \texttt{<second last product>}. \\
How likely are you to purchase \texttt{<last product>}? \\[7px]

A. Yes: The product logically and reasonably matches the characteristics I value. \\
B. Maybe yes: I might consider this product, but it doesn't strongly appeal to me. \\
C. Maybe no: The product does not directly relate to the characteristic I value. \\
D. No: I would not purchase this product if I were focused on the given characteristic. \\[7px]

Your Answer (Answer A or B or C or D only):
\end{tabular}\\ 
\midrule
\textsc{Task 3} & \begin{tabular}[c]{@{}l@{}}
Act as a customer who is browsing a series of products given as follows. \\
\texttt{<session product information>} \\[7px]

Comparing between \texttt{<last two products>}, \\
and assuming you have the intention of \texttt{<last two intention>}, \\
Does this comparison \texttt{<last intention comparison>} provide an in-depth justification of your impulse? \\[7px]

A. Yes: the comparison is reasonable and detailed enough to justify the change. \\
B. Maybe yes: The comparison could be more detailed and thorough but can be ignored. \\
C. Maybe no: The comparison is not entirely reasonable or lacks sufficient in-depth detail. \\
D. No: The comparison does not provide any underlying reasons or insights. \\[7px]

Your Answer (Answer A or B or C or D only):
\end{tabular}\\ 
\midrule
\textsc{Task 4} & \begin{tabular}[c]{@{}l@{}}
Act as a customer who is browsing a series of products given as follows. \\
\texttt{<session product information>} \\[7px]

After seeing \texttt{<previous products>}, \\
and assuming you have the intention of \texttt{<previous intention>}, \\
do you still want to explore similar products? \\[7px]

A. Yes: I want to explore products under the same category. \\
B. Maybe yes: I want to explore products under the same category but with different features. \\
C. No: I want to explore products under other categories. \\[7px]

Your Answer (Answer A or B or C only):
\end{tabular}\\ 
\bottomrule
\end{tabular}
}
\caption{Zero-shot prompts for model evaluation. \textsc{Task 1} stands for \textbf{\textit{\taskI{}}}, \textsc{Task 2} stands for \textbf{\textit{\taskII{}}}, \textsc{Task 3} stands for \textbf{\textit{\taskIII{}}}, \textsc{Task 4} stands for \textbf{\textit{\taskIV{}}}.
}
\label{tab:appendix_eval_zeroshot_prompts}
\end{table*}


\begin{table*}[t]
\small
\centering
\resizebox{\linewidth}{!}{
\begin{tabular}{@{}l|l@{}}
\toprule
Task & 5-shots Prompt \\ 
\midrule
\textsc{Task 1} & \begin{tabular}[c]{@{}l@{}}
Act as a customer who is browsing a series of products given as follows. \\
\texttt{<session product information>} \\
You hold an assumed intention, which will be provided later. \\
After seeing the products, you will be asked to determine the likelihood of purchasing the last product $\ \backslash$\\
based on the assumed intention. \\
You will be given four options to choose from: Yes, Maybe yes, Maybe no, No. \\
Please select the most appropriate option based on the given context. \\
A. Yes: The product is a logical and reasonable outcome of the purchasing intention. \\
B. Maybe yes: I may consider this, but it's not a strong impulse. \\
C. Maybe no: The product is not directly related to my intention. \\
D. No: I would never purchase it if I were the customer with the given intention. \\[7px]

Here are a few examples: \\
Q: After seeing Eco-friendly laundry detergent, bamboo dish brush, reusable kitchen cloths, \\
and assuming you are a customer who have the intention of Reducing household chemical usage. \\
How likely are you to purchase A biodegradable dish soap based on the assumed intention? \\
A: A. Yes \\[7px]

Q: After seeing Instant Pot, KitchenAid Stand Mixer, Ninja Air Fryer, \\
and assuming you are a customer who have the intention of Upgrading kitchen equipment for home cooking. \\
How likely are you to purchase A set of gourmet spices based on the assumed intention? \\
A: C. Maybe no \\[7px]

Q: After seeing Columbia hiking boots, North Face backpack, Garmin GPS watch, \\
and assuming you are a customer who have the intention of Planning for outdoor adventures. \\
How likely are you to purchase A formal suit for weddings based on the assumed intention? \\
A: D. No \\[7px]

Q: After seeing "1984" by George Orwell, "To Kill a Mockingbird" by Harper Lee, $\ \backslash$\\ "The Catcher in the Rye" by J.D. Salinger, \\ 
and assuming you are a customer who have the intention of Finding new reading material for leisure. \\
How likely are you to purchase "The Da Vinci Code" by Dan Brown based on the assumed intention? \\
A: B. Maybe yes \\[7px]

Q: After seeing Rolex Submariner, Omega Seamaster, Tag Heuer Monaco, \\
and assuming you are a customer who have the intention of Finding a timeless gift for a special occasion. \\
How likely are you to purchase A limited edition Patek Philippe watch based on the assumed intention? \\
A: A. Yes \\[7px]

Q: After seeing \texttt{<previous products>}, \\
and assuming you are a customer who have the intention of \texttt{<second last intention>}. \\
How likely are you to purchase \texttt{<last product>} based on the assumed intention? \\
A: \\
\end{tabular}\\  
\bottomrule
\end{tabular}
}
\caption{5-shots prompts for model evaluation. \textsc{Task 1} stands for \textbf{\textit{\taskI{}}}
}
\label{tab:appendix_eval_5shots_prompts_task1}
\end{table*}


\begin{table*}[t]
\small
\centering
\resizebox{\linewidth}{!}{
\begin{tabular}{@{}l|l@{}}
\toprule
Task & 5-shots Prompt \\ 
\midrule
\textsc{Task 2} & \begin{tabular}[c]{@{}l@{}}
Act as a customer who is browsing a series of products given as follows. \\
\texttt{<session product information>} \\
You have a valued feature/attribute, which will be provided later. \\
After seeing the products, you will be asked to determine the likelihood of purchasing the last product $\ \backslash$\\ based on the valued attribute. \\
You will be given four options to choose from: Yes, Maybe yes, Maybe no, No. \\
Please select the most appropriate option based on the given context. \\
A. Yes: The product logically and reasonably matches the characteristics I value. \\
B. Maybe yes: I might consider this product, but it doesn't strongly appeal to me. \\
C. Maybe no: The product does not directly relate to the characteristic I value. \\
D. No: I would not purchase this product if I were focused on the given characteristic. \\[7px]

Here are a few examples: \\
Q: After seeing Noise-canceling headphones, wireless earbuds, Bluetooth speaker, \\
and assuming you are a customer who highly value the feature High audio quality of Bluetooth speaker. \\
How likely are you to purchase A premium soundbar? \\
A: A. Yes \\[7px]

Q: After seeing adjustable standing desk, monitor with blue light filter, Ergonomic office chair, \\
and assuming you are a customer who highly value the feature Ergonomics of Ergonomic office chair. \\
How likely are you to purchase A desk lamp with a USB port? \\
A: C. Maybe no \\[7px]

Q: After seeing Organic facial cleanser, natural moisturizer, chemical-free sunscreen, \\
and assuming you are a customer who highly value the feature Natural ingredients of chemical-free sunscreen. \\
How likely are you to purchase A synthetic fragrance? \\
A: D. No \\[7px]

Q: After seeing DSLR camera, camera tripod, external flash, \\
and assuming you are a customer who highly value the feature Professional photography of external flash. \\
How likely are you to purchase A photo editing software? \\
A: A. Yes \\[7px]

Q: After seeing High SPF sunscreen, UV-blocking sunglasses, wide-brimmed hat, \\
and assuming you are a customer who highly value the feature Sun protection of wide-brimmed hat. \\
How likely are you to purchase An aloe vera gel? \\
A: B. Maybe yes \\[7px]

Q: After seeing \texttt{<previous products>}, \\
and assuming you are a customer who highly value the feature \texttt{<second last intention attribute>} $\ \backslash$ \\ of \texttt{<second last product>}. \\
How likely are you to purchase \texttt{<last product>}? \\
A: \\
\end{tabular}\\  
\bottomrule
\end{tabular}
}
\caption{5-shots prompts for model evaluation. \textsc{Task 2} stands for \textbf{\textit{\taskII{}}}.
}
\label{tab:appendix_eval_5shots_prompts_task2}
\end{table*}


\begin{table*}[t]
\small
\centering
\resizebox{\linewidth}{!}{
\begin{tabular}{@{}l|l@{}}
\toprule
Task & 5-shots Prompt \\ 
\midrule
\textsc{Task 3} & \begin{tabular}[c]{@{}l@{}}
Act as a customer who is browsing a series of products given as follows. \\
\texttt{<session product information>} \\
You have an assumed intention, which will be provided later. \\
You will be asked to evaluate the provided comparison between the last two products  $\ \backslash$\\ based on the assumed intention. \\
You will be given four options to choose from: Yes, Maybe yes, Maybe no, No. \\
Please select the most appropriate option based on the given context. \\
A. Yes: the comparison is reasonable and detailed enough to justify the change. \\
B. Maybe yes: The comparison could be more detailed and thorough but can be ignored. \\
C. Maybe no: The comparison is not entirely reasonable or lacks sufficient in-depth detail. \\
D. No: The comparison does not provide any underlying reasons or insights. \\[7px]

Here are a few examples: \\
Q: Comparing between a budget smartphone with a long battery life and A high-end smartphone with $\ \backslash$\\ superior low-light performance, \\
and assuming you have the intention of Finding a device with the best camera quality, \\
Does this comparison The high-end smartphone boasts advanced camera technology  $\ \backslash$\\ provide in-depth justification of your impulse? \\
A: A. Yes \\[7px]

Q: Comparing between A compact car and a mid-size SUV, \\
and assuming you have the intention of Prioritizing fuel efficiency, \\
Does this comparison the mid-size SUV, although spacious, consumes more fuel due to its larger engine  $\ \backslash$\\ and heavier body provide in-depth justification of your impulse? \\
A: B. Maybe yes \\[7px]

Q: Comparing between A luxury wristwatch and a fitness tracker, \\
and assuming you have the intention of Tracking health metrics, \\
Does this comparison Finding a more affordable watch provide in-depth justification of your impulse? \\
A: D. No \\[7px]

Q: Comparing between A leather office chair with plush cushioning and $\ \backslash$\\ a mesh office chair with lumbar support \\
and assuming you have the intention of Seeking maximum comfort during long working hours, \\
Does this comparison The mesh office chair offers better breathability and ergonomic support  $\ \backslash$\\ provide in-depth justification of your impulse? \\
A: A. Yes \\[7px]

Q: Comparing between A hardcover book and an e-reader, \\
and assuming you have the intention of Enhancing the reading experience, \\
Does this comparison The hardcover book provides a tactile, while the e-reader offers portability, $\ \backslash$\\adjustable text size provide in-depth justification of your impulse? \\
A: C. Maybe no \\[7px]

Q: Comparing between \texttt{<last two products>}, \\
and assuming you have the intention of \texttt{<last two intention>}, \\
Does this comparison \texttt{<last intention comparison>} provide in-depth justification of your impulse? \\
A: \\
\end{tabular}\\  
\bottomrule
\end{tabular}
}
\caption{5-shots prompts for model evaluation. \textsc{Task 3} stands for \textbf{\textit{\taskIII{}}}.
}
\label{tab:appendix_eval_5shots_prompts_task3}
\end{table*}


\begin{table*}[t]
\small
\centering
\resizebox{0.9\linewidth}{!}{
\begin{tabular}{@{}l|l@{}}
\toprule
Task & 5-shots Prompt \\ 
\midrule
\textsc{Task 4} & \begin{tabular}[c]{@{}l@{}}
Act as a customer who is browsing a series of products given as follows. \\
\texttt{<session product information>} \\
You will be provided with a sequence of intention. \\
You will be asked to determine whether you still want to explore similar products $\ \backslash$\\based on the sequence of intention. \\
You will be given three options to choose from: Yes, Maybe yes, No. \\
Please select the most appropriate option based on the given context. \\
A. Yes: I want to explore products under the same category. \\
B. Maybe yes: I want to explore products under the same category but with different features. \\
C. No: I want to explore products under other categories. \\[7px]

Here are a few examples: \\
Q: After seeing Stainless steel kitchen knives, non-stick frying pans, silicone spatulas, \\
and assuming you have the intention of Upgrading kitchen tools for home cooking,\\
do you still want to explore similar products? \\
A: B. Maybe yes \\[7px]

Q: After seeing Fitness tracker, yoga mat, resistance bands, \\
and assuming you have the intention of Tracking fitness progress, \\
do you still want to explore similar products? \\
A: A. Yes \\[7px]

Q: After seeing Stainless steel refrigerator, smart oven, induction cooktop, \\
and assuming you have the intention of Making the kitchen more energy efficient, \\
do you still want to explore similar products? \\
A: C. No \\[7px]

Q: After seeing Smart thermostat, LED light bulbs, energy-efficient washing machine, \\
and assuming you have the intention of Saving on utility bills, \\
do you still want to explore similar products? \\
A: B. Maybe yes \\[7px]

Q: After seeing Indoor plants, plant stands, watering can, \\
and assuming you have the intention of Creating a greener living space, \\
do you still want to explore similar products? \\
A: A. Yes \\[7px]

Q: After seeing \texttt{<previous products>}, \\
and assuming you have the intention of \texttt{<previous new intention>}, \\
do you still want to explore similar products? \\
A: \\
\end{tabular}\\  
\bottomrule
\end{tabular}
}
\caption{5-shots prompts for model evaluation. \textsc{Task 4} stands for \textbf{\textit{\taskIV{}}}.
}
\label{tab:appendix_eval_5shots_prompts_task4}
\end{table*}


\begin{table*}[t]
\small
\centering
\resizebox{\linewidth}{!}{
\begin{tabular}{@{}l|l@{}}
\toprule
Task & Chain-of-Thought Prompt \\ 
\midrule
\textsc{Task 1} & \begin{tabular}[c]{@{}l@{}}
Act as a customer who is browsing a series of products given as follows. \\
\texttt{<session product information>} \\[7px]

After seeing \texttt{<previous products>}, \\
and assuming you are a customer who have the intention of \texttt{<second last intention>}. \\
How likely are you to purchase \texttt{<last product>} based on the assumed intention? \\[7px]

A. Yes: The product is a logical and reasonable outcome of the purchasing intention. \\
B. Maybe yes: I may consider this, but it's not a strong impulse. \\
C. Maybe no: The product is not directly related to my intention. \\
D. No: I would never purchase it if I were the customer with the given intention. \\[7px]

Answer with a brief rationale then make your final choice $\ \backslash$\\by answering the option alphabet A/B/C/D only in the last line of your response. \\
Your Answer: \\
\end{tabular}\\ 
\midrule
\textsc{Task 2} & \begin{tabular}[c]{@{}l@{}}
Act as a customer who is browsing a series of products given as follows. \\
\texttt{<session product information>} \\[7px]

After seeing \texttt{<previous products>}, \\
and assuming you are a customer who highly value the feature  \texttt{<second last intention attribute>} $\ \backslash$\\ of \texttt{<second last product>}. \\
How likely are you to purchase \texttt{<last product>}? \\[7px]

A. Yes: The product logically and reasonably matches the characteristic I value. \\
B. Maybe yes: I might consider this product, but it doesn't strongly appeal to me. \\
C. Maybe no: The product does not directly relate to the characteristic I value. \\
D. No: I would not purchase this product if I were focused on the given characteristic. \\[7px]

Answer with a brief rationale then make your final choice $\ \backslash$\\by answering the option alphabet A/B/C/D only in the last line of your response. \\
Your Answer: \\
\end{tabular}\\ 
\midrule
\textsc{Task 3} & \begin{tabular}[c]{@{}l@{}}
Act as a customer who is browsing a series of products given as follows. \\
\texttt{<session product information>} \\[7px]

Comparing between \texttt{<last two products>}, \\
and assuming you have the intention of \texttt{<last two new intention>}, \\
Does this comparison \texttt{<last intention comparison>} provide in-depth justification of your impulse? \\[7px]

A. Yes: the comparison is reasonable and detailed enough to justify the change. \\
B. Maybe yes: The comparison could be more detailed and thorough but can be ignored. \\
C. Maybe no: The comparison is not entirely reasonable or lacks sufficient in-depth detail. \\
D. No: The comparison does not provide any underlying reasons or insights. \\[7px]

Answer with a brief rationale then make your final choice $\ \backslash$\\by answering the option alphabet A/B/C/D only in the last line of your response. \\
Your Answer: \\
\end{tabular}\\ 
\midrule
\textsc{Task 4} & \begin{tabular}[c]{@{}l@{}}
Act as a customer who is browsing a series of products given as follows. \\
\texttt{<session product information>} \\[7px]

After seeing \texttt{<previous products>}, \\
and assuming you have the intention of \texttt{<previous new intention>}, \\
do you still want to explore similar products? \\[7px]

A. Yes: I want to explore products under the same category. \\
B. Maybe yes: I want to explore products under the same category but with different features. \\
C. No: I want to explore products under other categories. \\[7px]

Answer with a brief rationale, then make your final choice $\ \backslash$\\by answering the option alphabet A/B/C only in the last line of your response. \\
Your Answer: \\
\end{tabular}\\ 
\bottomrule
\end{tabular}
}
\caption{Chain-of-Thought prompts for model evaluation. \textsc{Task 1} stands for \textbf{\textit{\taskI{}}}, \textsc{Task 2} stands for \textbf{\textit{\taskII{}}}, \textsc{Task 3} stands for \textbf{\textit{\taskIII{}}}, \textsc{Task 4} stands for \textbf{\textit{\taskIV{}}}.
}
\label{tab:appendix_eval_CoT_prompts}
\end{table*}

\end{document}